\documentclass[conference]{IEEEtran}
\IEEEoverridecommandlockouts

\usepackage{cite}
\usepackage{amsmath,amssymb,amsfonts}
\usepackage{subcaption}
\usepackage{algorithmic}
\usepackage{algorithm2e}
\RestyleAlgo{ruled} 
\usepackage{graphicx}
\usepackage{textcomp}
\usepackage{xcolor}
\newtheorem{definition}{Definition}
\newtheorem{theorem}{Theorem}[section]
\newtheorem{lemma}[theorem]{Lemma}
\def\BibTeX{{\rm B\kern-.05em{\sc i\kern-.025em b}\kern-.08em
    T\kern-.1667em\lower.7ex\hbox{E}\kern-.125emX}}
\newcommand{\yehuda}[1]{{ #1}}
\newcommand{\remove}[1]{{}}

\begin{document}

\title{Explainable Disparity Compensation for Efficient Fair Ranking

\thanks{© 2024 IEEE. Personal use of this material is permitted. Permission from IEEE must be
obtained for all other uses, in any current or future media, including
reprinting/republishing this material for advertising or promotional purposes, creating new
collective works, for resale or redistribution to servers or lists, or reuse of any copyrighted
component of this work in other works}
}

\author{\IEEEauthorblockN{Abraham Gale}
\IEEEauthorblockA{\textit{Department of Computer Science} \\
\textit{Rutgers, the State University of New Jersey}\\
Piscataway, USA \\
abraham.gale@rutgers.edu}
\and
\IEEEauthorblockN{Am\'elie Marian}
\IEEEauthorblockA{\textit{Department of Computer Science} \\
\textit{Rutgers, the State University of New Jersey}\\
Piscataway, USA \\
amelie.marian@rutgers.edu}
}
\maketitle
\IEEEpubidadjcol
\begin{abstract}
Ranking functions that are used in decision systems often produce disparate results for different populations because of bias in the underlying data. Addressing, and compensating for, these disparate outcomes is a critical problem for fair decision-making. Recent compensatory measures have mostly focused on opaque transformations of the ranking functions to satisfy fairness guarantees or on the use of quotas or set-asides to guarantee a minimum number of positive outcomes to members of underrepresented groups. 
In this paper we propose easily explainable data-driven compensatory measures for ranking functions. Our measures rely on the generation of bonus points given to members of underrepresented groups to address disparity in the ranking function. The bonus points can be set in advance, and can be combined, allowing for considering the intersections of representations and giving better transparency to stakeholders. We propose efficient sampling-based algorithms to calculate the number of bonus points to minimize disparity. We validate our algorithms using real-world school admissions and recidivism datasets, and compare our results with that of existing fair ranking algorithms.
\end{abstract}

\section{Introduction}

Ranking functions are used in a wide range of decision-making systems, such as resource allocation, candidate selection, or risk assessment. These ranking mechanisms often produce disparate outcomes because of bias in the underlying data. Compensating for disparities is therefore a key task for modern decision-making systems to ensure fairer, more equitable outcomes. Understanding the source of bias, which is sometimes not obvious but  hidden in correlations, is critical to address the disparity in the results. In this paper, we propose data-driven disparity compensation measures to transparently adjust ranking mechanisms based on the underlying data. Our measures are designed to be easily explainable to stakeholders in order to ensure  accountable and trustworthy decision-making systems.

Disparity compensation mechanisms for ranking functions used in real-world systems have mostly relied on the use of quotas, soft or hard, to ensure a minimum representation of members of protected groups. Quotas (or set-asides) have the advantage of being simple to implement and explain when only one dimension of inequity is present (e.g., set-asides for low-income students for school admissions). However, once several dimensions of disparity need to be accounted for (e.g, English language learners, low-income students, students with disabilities), the use of quotas becomes cumbersome and difficult to implement: Do members of two or more protected classes count towards one or more quotas? Does every combination of protected classes get a separate quota?~\cite{sonmez2019affirmative} As more dimensions are added, setting accurate set-aside thresholds can seem arbitrary and capricious to stakeholders.

We propose a mechanism based on the use of compensatory bonus points to address disparity, as defined ~\cite{Gale2020ExplainingMR} (see Section~ \ref{sec:disparity}), in ranking applications. Disparity represents the lack of statistical parity and is computed by measuring the distance between the selected (high-rank) and unselected (low-rank) objects in the fairness attribute space. Our bonus points approach is simple and easily understandable, as it links bonus points directly to sources of bias in the data. It allows for composing bonus points to model the intersectionality of bias and the compounding effect of different sources of disparity on ranking decision outcomes. It can be quickly and easily adjusted to new data and scenarios. 

We present algorithms for identifying the bonus points values that will minimize disparate outcomes on a given data distribution. Our work is based on a sample-based approach, which considers the underlying data distribution and draws samples to calculate the optimal number of bonus points to allocate to each disparity factor. This sample-based approach has three main benefits over the state of the art. First, it can be used to identify a set of compensatory bonus points before all the data is gathered, as long as the underlying distribution is known, by generating samples over the expected distribution. This can be beneficial in applications where providing transparent information to stakeholders is critical (e.g., letting students know in advance on which criteria they will be ranked, which equity-focused adjustments are in place, and how that would affect them). Second, by processing small samples of the data rather than the whole data set, we are able to identify high-quality compensatory bonus points in sub-linear time, compared to existing techniques which run in as high as exponential runtimes and are unpractical for large datasets. This makes our approach usable in many real-world use cases. Third, using bonus points is flexible enough to account for multiple sources of bias and disparity by allowing for the compounding effects of points to compensate for multiple disparate impacts.

We make the following contributions:
\begin{itemize}
    \item A model of disparity compensation for ranking functions based on the attribution of bonus points to members of protected classes. (Section~\ref{sec:background})
    \item A  disparity compensation algorithm (DCA) to identify the optimal value of compensatory bonus points to minimize disparity for a top-$k$ selection set. DCA runs in sub-linear time (Section~\ref{sec:algo}). We propose a modification of DCA that takes into account the whole ranking, using logarithmic discounting techniques~\cite{yang2017measuring}, to adapt to different selection sizes (Section~\ref{sec:log_discount}).

    \item An extensive evaluation of the impact of our compensatory mechanism over two real-world data sources: NYC public school student records used for high school admissions, and a dataset used to estimate recidivism risk in bail and sentencing decisions. 
    (Sections~\ref{sec:settings} and~\ref{sec:experiments})
    \item A comparison with state-of-the-art fair-ranking techniques that show that our proposed DCA algorithm results in comparable or better disparity reduction outcomes while being significantly more efficient. (Section~\ref{sec:Comparison}) 
\end{itemize}
We present related work in Section~\ref{sec:related}, present motivating examples in Section~\ref{sec:background}, and conclude in Section~\ref{sec:conclusion}.

\section{Related Work}
\label{sec:related}

The problem of providing ranking functions and systems with fair outcomes has recently received considerable scrutiny.
Several surveys provide a good overview of the literature both in the Data Management community~\cite{zehlike2021fairness,Fair_Classification}, and in the Recommender Systems community \cite{pitoura2021fairness}. These surveys classify fairness ranking approaches into  pre-processing, in-processing and post-processing techniques depending on when the fairness metrics are applied. Pre-processing techniques transform data before applying the ranking function~\cite{lahoti2019ifair}. Our approach could be classified as pre-processing as the bonus points are applied to the data before ranking.

Several types of pre-processing techniques have been discussed for use by classifiers. Their goal is to pre-process data so that an arbitrary classifier will be less likely to result in an unfair outcome. These transformations are usually done without prior knowledge of the classifier mechanics which can limit this approach's effectiveness. A seminal paper ~\cite{kamiran2012data} discusses several basic techniques for pre-processing data for classifiers, such as suppression, where some attributes of the objects are removed entirely, and sampling, where some objects are duplicated and others are removed. They also discuss a fairness-accuracy trade-off: as the fairness of the data increases the classifiers become less accurate; this tradeoff is further explored in~\cite{feldman2015certifying}, which provides provably fair solutions, and ~\cite{calmon2017optimized}, which gives a solution based on convex optimization. Similarly,  \cite{salimi2019interventional} and \cite{lahoti2019ifair} address a similar problem  using different notions of fairness:  causal fairness, where the user can specify which attributes are allowed to influence the result~\cite{salimi2019interventional}, and individual fairness where the goal is to make sure that objects are treated similarly to their nearest neighbors~\cite{lahoti2019ifair}.

For ranking applications, in-processing approaches are  more commonly used. These approaches adjust the ranking function to optimize a given fairness goal, either through learning techniques~\cite{radlinski2008learning} or manipulation of weighted functions~\cite{asudeh2019designing}. Celis et al.~\cite{celis2017ranking} propose approximation algorithms to provide rankings as close to the original rank quality metric while satisfying fairness constraints. Yang and Stoyanovich~\cite{yang2017measuring} consider a model of probability-based fairness where each class of objects should be treated equally. This probabilistic interpretation is also considered in~\cite{zehlike2017fa}, which provides probabilistic fairness guarantees in settings with only one protected class; this setting is relaxed in~\cite{zehlike2022fair}, but the approach relies on a Cartesian product of all protected classes and is prohibitively expensive. A drawback of these approaches is that they result in ranking functions that are often opaque and hard to explain to stakeholders; in contrast, we aim to make the disparity compensation process transparent and easily explainable.

Several notions of fairness, as it relates to rankings,  have been proposed; \cite{dwork2012fairness} discusses the notion of group fairness vs. individual fairness in the context of fair classification and argues that  statistical parity is a desirable  measure of fairness. The Disparity metric that we use (Section~\ref{sec:definitions}) can be interpreted as a measure of statistical parity~\cite{Gale2020ExplainingMR}. Exposure~\cite{singh2018fairness} is often used as a measure of fairness, although the definition of exposure varies in the literature; we use the same definition as~\cite{gupta2021online} in our experimental evaluation (Section~\ref{sec:exposure}). Fair ranking inherently involves choices. In \cite{kleinberg2016inherent} the authors prove that different fairness metrics are incompatible, even in the approximate case. Before any dataset-specific choices are made, a choice of how to define fairness for the given problem is required. As mentioned above, we focus on a statistical parity interpretation of fairness, as it is easily explainable and does not require making assumptions about underlying sources of bias.


A common real-world application of fair rankings is school admissions. School districts have used lotteries to select students as a way to integrate schools and reduce bias in selection outcomes. However, lotteries are not a simple answer to solve for disparities and can in fact exacerbate selection bias~\cite{nycd1} or achievement bias~\cite{collegelotteries2021}. To ensure diversity, many school districts such as NYC and Chicago~\cite{ellison2021efficiency} have successfully opted for set-aside quotas for low-income students. However, these affirmative action-based approaches have been shown to be potentially detrimental to minorities~\cite{ehlers2014,kojima2012,hafalir2013} depending on the underlying school choice patterns. In addition, quota-based approaches are difficult to implement when multiple groups should be given protected treatment. Works have considered overlapping quotas~\cite{sonmez2019affirmative} to account for a candidate exhibiting multiple protected characteristics, minimum and maximum bounds for quotas~\cite{aziz2021multi}, or priority assignments rules for quotas~\cite{abdulkadirouglu2021priority}. These approaches are all computationally expensive. In contrast, our disparity compensation bonus points allow for (1) targeted intervention for each dimension of disparity, even overlapping ones, and (2) a computationally efficient approach.

\section{Background}
\label{sec:background}
Ranking functions are used in a wide variety of decision systems with high societal impacts: job recruiting tools, school admissions, allocation of resources (e.g., vaccines, treatments, public housing), or risk assessment (e.g., fraud, recidivism). Ensuring that these mechanisms are fair is critical. To this end, we propose a model of disparity compensation measures based on the allocation of targeted bonus points.

We introduce two motivating examples for our work in Section~\ref{sec:motivations}, and set the definitions and parameters of our problem in Section~\ref{sec:definitions}.

\subsection{Motivating Examples}
\label{sec:motivations}

\paragraph{\textbf{NYC High School Admissions}} NYC high school admissions use a deferred acceptance (DA) matching
algorithm~\cite{NYCmatching} similar to the stable marriage algorithm designed by Gale-Shapley~\cite{galeshapley}. The  algorithm  matches students to schools based on students'
preferences and the schools' admission-ranked lists (rubrics). Schools set their own ranking rubrics using metrics such as grades, test scores,  
absences, auditions, or interviews.  Such screens have become a topic of controversy, being targeted as discriminatory because
the underlying metrics often exhibit a high level of
disproportionality in the ranked lists they produce, as students from disadvantaged groups often score lower in some of the metrics used in the rubrics.  

Currently, the NYC Department of Education mostly relies on set-asides (soft quotas) for low-income students to address disparity and produce a diverse group of students at each school. These measures have had mixed results depending on the demographics of the geographical area of the school and on the patterns of student choices. These set-asides are mostly limited to low-income students, although other dimensions of disadvantage have been considered (current school, English language learner, student in temporary housing). However, as mentioned in Section~\ref{sec:related}, combining several quotas can be cumbersome and computationally expensive.

In this paper, we explore the use of compensatory “bonus points" assigned to students who exhibit one or more
dimensions of disadvantage. Our goal is to ensure adequate representation of the underlying population in the students selected by the school
admission rubrics. Because NYC uses a matching algorithm, it is not known in advance how far down its list a school
will accept students; our techniques can adjust to unknown values of the number of selected objects $k$ by minimizing the disparity over all values of $k$, logarithmically discounted to favor smaller $k$ values (Section~\ref{sec:log_discount}). Our experimental evaluation, using NYC high school admission data, shows that our compensatory metrics adapt well to multiple selection percentages (Section~\ref{sec:resschools}).

Some school systems have considered the use of a point-based scheme to diversify schools. In particular, Paris, France has shown good results in improving
socioeconomic diversity~\cite{affelnet} through the use of ``bonus" points for disadvantaged students. However, the system was based on ad hoc bonus points decided somewhat arbitrarily by policymakers, which created some undesirable outcomes in some schools when the points were not calibrated correctly: in one case a high school was assigned a large majority (83\%) of low-income students (instead of the statistical parity goal of 40\%), defeating the diversity purpose. In contrast, we propose a data-driven assignment of bonus points that best reflects the data distribution and its impact on the rankings.

\paragraph{\textbf{COMPAS Recidivism Data}}
Recidivism algorithms, such as COMPAS, are used to predict the likelihood that a person interacting with the criminal justice system will re-offend, and are used by U.S. Courts to  assist in bail and sentencing decisions. The impact of these decision algorithms  is unquestionable, yet the opacity of the decisions makes it hard to verify that the process is fair.
 
 In 2016 a ProPublica investigation~\cite{angwin_larson_2016} argued that the COMPAS algorithm was unfair to a number of disadvantaged groups, particularly Black Americans. The internal COMPAS ranking algorithm has not been made public; ProPublica based its investigation on  Broward County, Florida data acquired through a public records request, which they made public. Subsequently, this COMPAS data has become a popular dataset for evaluating fairness mechanisms; according to the fairness survey in~\cite{zehlike2021fairness}, it is the most popular large dataset for fairness analysis.

While COMPAS is used for classifying subjects into categories, the categories (deciles) are based on an underlying ranking of subjects. The deciles scores are then often (mistakenly) used as absolute, and not relative, scores of recidivism. Because the scores are based on comparative data, they exacerbate underlying discriminatory practices.
The internal COMPAS algorithm is proprietary, its inner workings, and potential disparate treatments, have been the subject of dispute and speculation~\cite{Rudin2020Age,Jackson2020Setting}. Yet the disparate impact of the COMPAS decile scores
, as they are used in practice, is undeniable. In addition, the process is opaque and not easily understandable. We explore using our disparity compensation techniques in conjunction with the COMPAS scores to address the disparate impacts of the COMPAS tool and report on our results in Section~\ref{sec:rescompas}.

The use of the COMPAS dataset has been the subject of multiple criticisms regarding the ethical use of such data~\cite{COMPASmessy21}.
Our inclusion of COMPAS as a case study is by no means an endorsement of its use for real-life decisions, but rather an illustration of how our compensatory-based approach can help significantly reduce disproportionality on various types of ranking –and classification– functions, even when those are hidden behind black-box proprietary systems.

\vspace{-0.05in}
\subsection{Definitions}
\label{sec:definitions}

We focus on explainability so that the fairness compensation choices are simple, transparent, and clearly understandable for stakeholders. This notion of
explainability is especially important to gain support from stakeholders~\cite{goeljustice}. We now define the format of our ranking functions (Section~\ref{sec:ranking functions}) and compensatory bonus points (Section~\ref{sec:bonuspoints}) and our choice of fairness metric (Section~\ref{sec:disparity}).

\subsubsection{Ranking Functions}
\label{sec:ranking functions}

We focus our explanation on score-based ranking functions. Our bonus points can also be adapted to other ranking functions by simulating an underlying score based on rank (see Section~\ref{sec:rescompas}).

\begin{definition}\textit{Score-Based ranking function}
\label{def:WeightedFunction}
We define a score-based ranking function $f$ over a set of $A$ attributes   $a_1,
...., a_A$, over an object $o$ as $f(o)=f(a_1,
...., a_A)$. A ranking process $R$  selects the $k$\% best objects with the highest $f(o)$ values as its answer $R_k$.

\end{definition}

Each object has a set of attributes $A$ that defines its properties and assigns values to them. For the purpose of this work, we recognize a special subset of attributes {\em fairness attributes} (also called protected attributes in the literature), which represent the dimensions on which we want to control for bias and disparate impact. {\em Fairness attributes} may be used by the ranking function $f$ to score the objects, or may not be involved in the ranking but still of interest for assessing the fairness of the outcome.

For example, a school may rank applicants using a 100-point scoring function based on a weighted sum of students' GPA and test scores (attributes). Fairness attributes may include low-income or disability status of the student.

\subsection{Bonus Points}
\label{sec:bonuspoints}

Our approach centers around bonus points to compensate for various dimensions of disparity. Bonus points are multiplied  with the corresponding fairness attribute value and added to the final ranking function score $f(o)$. When the fairness attribute value is binary, this is equivalent to adding the bonus to the final score when that value is equal to 1. For instance, if the low-income status of a school applicant is encoded as a \{0,1\} binary, a bonus of 2 points would add 2 to each low-income applicant's final score; if the low-income status is encoded as a continuous value in [0,1], then the bonus of 2 will be  multiplied by the value of the attribute to give a more precise disparity compensation tool.

We define bonus points as:
\begin{definition}\textit{Bonus Points}
\label{def:Bonus_points}
Given a vector of fairness attributes $\Vec{A_f}$ and a identically shaped vector of bonus points $\Vec{B}$ let the score of an object $o$ be defined as $f_b(o) = f(o) + \Vec{A_f} \cdot \Vec{B} $
\end{definition}

We require bonus points to be positive (negative for scenarios where a lower score is desirable). Negative bonus points would be perceived as a penalty and may not be easily accepted by stakeholders.

In addition to the flexibility of the mechanism, the advantages of using bonus points include: \textbf{intersectionality}, bonus points can be combined and compounded to account for multiple dimensions of bias; \textbf{transparency}, the extent and impact of the fairness intervention are clear to stakeholders; \textbf{comparability}, the score of objects can be easily adjusted and objects compared, increasing transparency and trust; \textbf{predictability}, combined with information on how the selection is done (e.g., historical threshold values), applicants can easily assess their chances and be provided with  \textbf{clarity} as to which actions or interventions are required for selection.

For instance, in our school admission scenario, bonus points could be used to capture the \textit{intersectionality} of students with disability and low-income students: students with both characteristics would receive more bonus points than students with one, or none. This information can \textit{transparently} be published before applications are due, giving clear and \textit{predictable}, and \textit{comparable}, information to families. Admission decisions are \textit{clarified}, with clear thresholds published and the participation of each ranking attribute and fairness compensatory bonus points identified for each applicant. \yehuda{This leads to a more transparent and explainable experience for the students, who know their score and fairness adjustments, if any, at the time of application.}


\subsection{Disparity}
\label{sec:disparity}

We focus on the explainable disparity  from~\cite{Gale2020ExplainingMR} as our target fairness metric as it aims at satisfying statistical parity~\cite{lahoti2019ifair}. Furthermore, it is easily interpretable by humans, behaves well even when the number of dimensions increases and can deal with dimensions with either continuous or discrete data.

Disparity is defined as the vector difference between the average selected object and the average unselected object. Formally it is defined as follows:
\begin{definition}\textit{Disparity}
\label{def:Disparity}
Given a set of $O$ objects and a selection $K$ of $k$ percent of objects in $O$,
Let $\vec{D}^F_O$ be the centroid of $O$ over a set of fairness attributes $F$, and let $\vec{D}^F_k$ be the centroid of the $K$ selected objects  over the same set of attributes. We define the disparity $\vec{D}^F$ as the $|F|$ dimensional disparity vector where $ \vec{D}^F \equiv \vec{D}^F_k - \vec{D}^F_O$.
\end{definition}
When the set of fairness attributes is understood, we omit $^F$ for simplicity of notation: $ \vec{D} \equiv \vec{D}_k - \vec{D}_O$

Intuitively, disparity measures the difference between the average selected object and the average object overall. \yehuda{The value of disparity for discrete dimensions represents the percentage of the selected set which would need to change to achieve statistical parity.} For example, if the population is 30\% low income and the selected set is 20\% low income that would lead to a disparity of \yehuda{30 - 20 = } 10\% disparity or 0.1. \yehuda{This means that if 10\% of the selected set is changed from non-low-income to low-income then there would be statistical parity.} For continuous fairness attributes, disparity is normalized based on the range of values. For instance, in a population with income in [\$0;\$200,000], if the average income of the population is \$40,000 (normalized to 0.2) and the average income of the selected set is \$100,000  (normalized to 0.5) that would lead to a disparity of \$60,000  (normalized to 0.3). Each fairness attribute is one dimension of the disparity vector. Assuming all values are normalized between 0 and 1, a disparity magnitude of -1 or 1 means that the protected attribute is either present only in the population, or only in the protected set respectively. A disparity of zero indicates statistical parity. 

\section{Disparity Compensation Methods}
\label{sec:algo}

We now present our disparity compensation approach. We first highlight several challenges we aim to address in Section~\ref{sec:challenges}. In Section~\ref{sec:DCA_algo} we describe the Disparity Compensation Algorithm (DCA), our primary contribution; we discuss the accuracy of DCA in Section~\ref{sec:accuracy}, and provide its time complexity  in Section~\ref{sec:timecomplexity}.

\subsection{Challenges}
\label{sec:challenges}
Our goal is to find the optimum number of points to allocate to protected groups in order to minimize disparity. This task can be thought of as an optimization task, in which the goal is to pick a bonus vector $\vec{B}$ such that the $L^2$ norm of the disparity vector $\vec{D}$ is minimized.  Formally, we want to minimize the disparity:
\begin{equation*}
\begin{aligned}
& \underset{B}{\text{minimize}}
& & ||\vec{D}(\vec{B})||_2 
& \text{subject to}
& & b_i \geq 0
\end{aligned}
\end{equation*}

Where $\Vec{D}(\vec{B})$ is the disparity on a given population as a function of the bonus vector as defined above, containing the number of bonus points given to each fairness attribute.
This minimization is complicated by the following challenges.

\begin{enumerate}
    \item There are a large number of possible solutions. This means that most traditional algorithmic solutions are very slow. For instance, recent fairness algorithms are super-polynomial~\cite{zehlike2022fair}, and not easily scalable.
    \item In a set selection task, such as identifying the $k$-objects, measures to assess the fairness of the rankings are step functions, as their value change with every new candidate selected, or each change in  the ranking order. A small change in the bonus point vector can therefore lead to an arbitrarily large change in the disparity, which  means the optimization functions are not smooth or continuous. As such, they are non-differentiable, and standard derivative-based optimization methods are inapplicable.  
    \item Our minimization function does not exhibit convexity, or even quasi-convexity, which precludes us from using convex optimization techniques.
    \item Evaluating each possible solution is expensive as it requires a re-ranking of the dataset. Non-differential (derivative-free) optimization solutions  are therefore inefficient because they typically re-rank the data hundreds of times ~\cite{belkhir2017per}.
    \item For practical purposes, it is desirable for our methods to be fast enough so that function designers (e.g. school administrators) can iterate over several options to assess the impacts of fairness adjustments. 
.
    
\end{enumerate}

To address these challenges, we propose using a novel descent-based method to compute the correct number of bonus points to eliminate disparity

\subsection{Disparity Compensation Algorithm}
\label{sec:DCA_algo}

\begin{algorithm}
\caption{Core DCA}
\SetAlgoLined
\KwResult{$\vec{B}$}
 O $\leftarrow$ the entire set of available objects\;
 k $\leftarrow$ size of the selection\;
 lr $\leftarrow$ list of learning rates sorted in decreasing order\;
 t $\leftarrow$ number of iterations\;
 $\vec{B}$ $\leftarrow$ weight vector of dimensionality equal to number of fairness attributes initialized randomly\;
\For{L in lr}{
  \For{x in t}{
  S $\leftarrow$ A random sample of $sample\_size$   from O\;
  $\vec{D}_k$ $\leftarrow$ Disparity of the $k$ selection over S after applying $\vec{B}$ bonus points\;
  $\vec{B}$ $\leftarrow$ $\vec{B} - L \times \vec{D}_k$\;
  \For{D in $\vec{B}$}{$D \leftarrow max(D, 0)$}
  
 }
}
\label{algo_SGD}
\end{algorithm}

\begin{algorithm}
\KwResult{$\vec{B}$}
 A $\leftarrow$ An array for computing the average\;
 O $\leftarrow$ the entire set of available objects\;
 $\vec{B}$ $\leftarrow$ The output vector of DCA\;
 k $\leftarrow$ size of the selection\;
 t $\leftarrow$ number of iterations for refinement\;
 \For{x in t}{
  S $\leftarrow$ The next sample in O\;
  $\vec{D}_k$ $\leftarrow$ Disparity with $\vec{B}$ bonus points over S\;
  $\vec{B}$ $\leftarrow$ Adam.step($\vec{B},\vec{D}_k$)\;
  A $\leftarrow$ $A + \vec{B}$\;
 }
 \Return ROUND(AVERAGE(A))\;
 \caption{DCA Refinement}
\label{refine_function}
\end{algorithm}

Traditional descent-based methods cannot be applied to our setting as the presence of  large plateaus and steps renders the function non-differentiable; standard gradient descent methods are derivative-based and cannot be used on non-differentiable optimization functions. We circumvent this issue by using the disparity vector directly, instead of its gradient.

Our Disparity Compensation Algorithm (DCA) (Algorithm~\ref{algo_SGD})  is based on the observation that any descent movement in a dimension that aims at compensating the disparity in that dimension will results in a better outcome as long as it does not flip the sign of the Disparity $\vec{D}$ (i.e., create a reverse disparate impact). 

We consider cases where we want to prevent disparate outcomes in future decisions, not only on a known dataset. Therefore, it is not enough to perfectly address the disparity of the training data, our solution has to be applicable to any similar dataset. The training data can then be seen as a sample drawn from an underlying distribution, and our goal is to minimize disparity for that distribution. Therefore we can use the Central Limit Theorem and the Quantile Central Limit Theorem~\cite{ruppert2011statistics} to estimate the selectivity of the ranking function on the underlying distributions (Section~\ref{sec:accuracy}). Our methods can be applied similarly in the absence of training data if the expected distribution of the dataset is known.

The algorithm is shown in Algorithm~\ref{algo_SGD}. DCA works by keeping a bonus vector, which is incrementally adjusted in the opposite direction of disparity. DCA loops though decreasing learning rates (step sizes) to reduce the disparity vector (noted $\vec{D}$) as close as possible to zero. For each learning rate, the algorithm adjusts the bonus vector over a fixed number of steps $t$ to get as close to zero as possible using that learning rate; it then goes down to the next learning rate.  At each step the Disparity is only computed on a small sample drawn uniformly at random from the overall distribution (or a representative training set). The entire set of objects $O$ is never looked at directly. For each learning rate, we take a fixed number of samples. In each step, DCA uses the Disparity on the sample to predict the Disparity on the distribution as a whole, using the current best guess for bonus points $\vec{B}$. DCA then adjusts the current best guess in the opposite direction of the disparity. For example, in the case above, if the population is 30\% low income and the selected set is 20\% low income that would lead to a 10\% disparity or 0.1. With a learning rate of 0.2, each Low-Income member of the population would receive $0.1 \times 0.2 = 0.02$ extra points in their scoring function for the next iteration. Then, a new sample would be taken and ranked and the disparity would be calculated again.

We propose a refinement step in Algorithm~\ref{refine_function}. The refinement consists of a for loop that uses an adaptive learning rate (using the Adam method~\cite{kingma2017adam}) to find the best estimate. Instead of using a fixed learning rate for all the parameters, the Adam method uses an individual learning rate for each parameter which is individually optimized based on the change in the gradient, or in our case the disparity. The Adam method is especially useful and popular to deal with the noise created by samples. 
Next, the average of the guesses is taken to further reduce the noise created by the random samples and get a more consistent result.  As we will see experimentally in Section~\ref{sec:refine}, this refinement step results in smoother Disparity compensation results.  

The values for $lr$ and $t$ provide a tradeoff between time and accuracy. We set them empirically for our experiments (Section~\ref{sec:settings}).

\subsection{Accuracy of DCA}
\label{sec:accuracy}
To show the accuracy of DCA, we first consider a variation of the DCA, called {\em Full DCA} that considers the entire dataset, not a sample. Consider two objects $p$ and $q$, $q$ in the top-$k$ and $p$ outside the top-$k$. If switching their positions will reduce the disparity, {\em Full DCA} will always reduce the difference in score between them. Formally:
\begin{theorem}
At every step of Full DCA, if removing object $q$ from the top-k and replacing it with object $p$ would reduce the overall disparity, Full DCA will allocate more bonus points at that step to $p$ than to $q$.
\end{theorem}
Mathematically this means that:
$$
(\vec{B} - L \times \vec{D}) \cdot \vec{F_q} - \vec{B} \cdot \vec{F_q} < (\vec{B} - L \times \vec{D}) \cdot \vec{F_p} - \vec{B} \cdot \vec{F_p})
$$
Or
$$0 > \vec{D}\cdot (\vec{F_p} - \vec{F_q})$$
Where $\vec{F_p}$ and $\vec{F_q}$ are the fairness attribute vectors, $\vec{D}$ is the disparity, $L$ is the step size, and $\vec{B}$ is the bonus vector. When using DCA without sampling this will always be true. This can be shown from the definition of disparity and the given assumption that switching these two objects will reduce disparity:
$$
|\vec{D}|_2 > |\frac{1}{s} \sum_{i \in \textbf{S}}F_i + \frac{1}{s}\times \vec{F_p} - \frac{1}{s}\times \vec{F_q} - \vec{Q}|_2
$$
Where Q is the centroid of the entire distribution (which is constant during the running of DCA) and s is the number of selected objects. Using the definition of the $L_2$ norm we see that the above inequality only holds if:
$$
\vec{D} \cdot \vec{D} > \vec{D} \cdot \vec{D} + 
\frac{2}{s}\times (\vec{F_p} - \vec{F_q}) \cdot \vec{D} + \frac{1}{s^2}\times (\vec{F_p} - \vec{F_q})\cdot(\vec{F_p} - \vec{F_q}))
$$
Which simplifies to:
$$
-\frac{1}{2s}\times (\vec{F_p} - \vec{F_q})\cdot(\vec{F_p} - \vec{F_q}) > \vec{D} \cdot (\vec{F_p} - \vec{F_q})  
$$
Since the left side is always negative, we have shown $$0 > \vec{D}\cdot (\vec{F_p} - \vec{F_q})$$ and that $p$ will always receive more additional bonus points than $q$.

Unlike {\em Full DCA}, DCA relies on samples of the distribution to efficiently identify the best bonus point vector $\vec{B}$ to apply on the set of fairness attributes $F$ to minimize disparity. The accuracy of DCA then depends on the accuracy of the computation of the Disparity metric $\vec{D}$ over the samples as  estimators of the Disparity over the whole dataset.

The Disparity $\vec{D}$ is computed as the distance between the centroid over the set of all objects $O$, $\vec{D}_O$, and the centroid of the $K$ selected objects, $\vec{D}_k$ (Section~\ref{sec:definitions}). In this section, we will show that computing $\vec{D}_O$ and $\vec{D}_k$ over a sample of the dataset gives a good estimation of their value over the whole dataset.

\begin{lemma}
The centroid $\vec{D_s}$ of a sample $s$ over a set of objects $O$ is an unbiased, low-error, estimate of the centroid $\vec{D}_O$ over the entire set of objects $O$.
\end{lemma}

This result directly follows from the Central Limit Theorem which states that the mean of a sample will approximate that of the original distribution as long as the sample size is sufficiently large (at least 30). 

Next, we show that the score of the object at the  $k^{th}$ percentile in the sample $s$ is a good estimator of the score of the object at the $k^{th}$ percentile over the whole distribution. 

\begin{lemma}
The score at quantile value $k$ of a sample $s$ over a set of objects $O$ is an unbiased, low-error, estimate of the score at quantile value $k$ over the entire set of objects $O$.
\label{lemma-quantile}
\end{lemma}

This lemma is a direct result of the Quantile Central Limit Theorem. \cite{ruppert2011statistics}, which says that the quantile $k$ of a sample approximates the corresponding quantile of the original distribution, as long as the density function of the sample at $1-k$ is positive. This qualification is met as long as the sample size is at least $\frac{1}{k}$, which gives us a lower bound on the sample size used in DCA. 

The variance of the sample quantile as an estimator can be large for values of $k$ close to 0 or 1, and  may lead to low-quality estimations in those cases. This is reflected in real-world experiments, as shown in section \ref{sec:varying}; however for  reasonable values of $k$, the $k^{th}$ percentile of the sample is an accurate estimator of the $k^{th}$ percentile of the distribution.

\begin{lemma}
The centroid $\vec{D_s}_k$ of the  $k$ percent selected objects over a sample $s$ over a set of objects $O$ is an unbiased, low-error, estimate of the centroid $\vec{D}_k$ of the  $k$ percent of selected objects over the entire set of objects $O$.
\end{lemma}

From Lemma~\ref{lemma-quantile} we know that the $k^{th}$ percentile value of the sample $s$ is an estimator of the $k^{th}$ percentile value of the distribution of all objects $O$. The top $k$ percent objects from $s$ are therefore taken from the same distribution as the top $k$ percent objects from $O$: the distribution of $O$ truncated at the $k^{th}$ percentile value. 

From the Central Limit Theorem, we know that the mean of a sample will approximate that of the original distribution as long as the sample size is sufficiently large (at least 30). Since both top $k$ percent selections over the sample $s$ and $O$ are taken from the same distribution, the Central Limit Theorem applies, and $\vec{D_s}_k$  is an unbiased, low-error, estimate of $\vec{D}_k$.

From the above results, it follows that:

\begin{theorem}
The sample Disparity $\vec{D_s} \equiv \vec{D_s}_k - \vec{D_s}_O$ is an unbiased, low-error, estimate of $ \vec{D} \equiv \vec{D}_k - \vec{D}_O$ of the Disparity over the whole dataset.
\end{theorem}

 \subsection{Time Complexity of DCA}
 \label{sec:timecomplexity}
 The time complexity of DCA does not depend on the size of the dataset but on the sample size and characteristics of the distribution, which allows for fast performance in practice. This is because DCA focuses on correcting the disparity in the underlying distribution rather than on a specific dataset. A training dataset represents a larger sample over the hidden distribution. This allows DCA's execution time to depend on the (smaller) samples' size but not on the dataset size. The algorithm takes a constant multiple of the time taken to compute the disparity on one sample. This time is $$O({sample\_size \times log(sample\_size)})$$ if the sample is fully sorted. 
 The size of the sample needs to be large enough so that the  Central Limit Theorem can be applied, this is generally recognized to be around 30. This means that $30 = sample\_size * k$ and the sample size is $O(\frac{1}{k})$. 
 
 In addition, each subgroup of interest needs to appear in the sample in a reasonable number, for the same reason, so the Central limit theorem can apply. This leads to a final sample size of $O(max(\frac{1}{k},\frac{1}{r}))$ where $k$ is the proportion of elements selected by the ranking process and $r$ is the frequency of the least common group in the dataset. Given this, assuming that $k$ is large enough that the entire sample must be sorted, the time complexity of DCA does not depend on the size of the dataset and is: $$O(max(\frac{1}{k},\frac{1}{r})\times log(max(\frac{1}{k},\frac{1}{r}))$$ 

\subsection{Adjusting the Optimization Goal of DCA for multiple values of $k$}
\label{sec:log_discount}

We have presented DCA for the case where the size of the selection $k$ is known in advance. However, it is often useful to optimize an entire ranking, either when the $k$ is unknown in advance (such as ranked lists in school matching applications), or when the ranking over the entire population is used. 

We propose a modification of DCA that updates the definition of disparity to use the whole ranking along with the logarithmic discounting techniques described in~\cite{yang2017measuring}, to assign more importance to objects selected first than to those selected last. Logarithmic discounting replaces the disparity at k with in our minimization goal of Section~\ref{sec:disparity} with:  $$\frac{1}{Z}\sum_{i\in10,20,30..}^{i=k}\frac{\vec{D}_i}{log_2(i + 1)}$$ 
Where Z is a normalization factor defined as the maximum possible value.

The computation of $\vec{D}_k$ in Algorithms~\ref{algo_SGD} and~\ref{refine_function} is  replaced with this new logarithmically discounted disparity. This new metric retains the useful characteristics of the previous one: it is a vector with each dimension representing an individual fairness attribute and calculated independently, it ranges between -1 for completely unfair in one direction to 1 for completely unfair in the other direction, is equal to 0 for fair representation, and it can  be summarized by its norm. 

 When using logarithmically-discounted disparity, the minimization problem of Section~\ref{sec:challenges} is then changed to:
 \begin{equation*}
\begin{aligned}
& \underset{B}{\text{minimize}}
& & \sum_{j\in10,20,30..}^{j=k}\frac{||\vec{D}_j(\vec{B})||_2}{log_2(j + 1)} 
& \text{subject to}
& & b_i \geq 0
\end{aligned}
\end{equation*}
Where $\vec{D}_j$ is the disparity with $k=j$.

In terms of time complexity, using the logarithmically-discounted disparity version of DCA takes longer by an additional factor of the size of the sample, as we need to evaluate disparity at every point in the sample, leading to an overall time of:
 $$O((max(\frac{1}{k},\frac{1}{r})\times log(max(\frac{1}{k},\frac{1}{r}))\times max(\frac{1}{k},\frac{1}{r}))$$

Often, only part of the ranking is interesting to the user. Logarithmic-discounted disparity can be adjusted to various ranking needs. For example, users might only be interested in the top half of the ranking. In this case, the disparity outside that section of the ranking can be ignored, and the discounted disparity can still be computed straightforwardly only for values of $k<\frac{N}{2}$.

\section{Experimental Setting}
\label{sec:settings}

We now describe the datasets used in our evaluation, the parameters we set for the implementation of our DCA algorithm, and our experimental environment.

\subsection{Datasets}

\paragraph{NYC school dataset}
We evaluate our algorithms using real student data from NYC, which we received through a NYC Data Request~\cite{nycdata}, and for which we have secured IRB approval.  

The data used in this paper consists of the grades, test scores, absences, and demographics of around 80,000 7$^{\text{th}}$ graders each for both the 2016-2017 and 2017-2018 academic years. NYC high schools use the admission matching system described in Section~\ref{sec:motivations} when students are in the 8$^{\text{th}}$ grade; the various attributes used for ranking students therefore are from their 7$^{\text{th}}$ grade report cards. We used data from the 2016-2017 academic year as our training data, and data from the 2017-2018 academic year as our test data. 

We selected our ranking function to model the admission function that several real NYC high schools used for admission in the years 2017 and 2018: a weighted-sum function $f=0.55* GPA + 0.45* TestScores$, where $GPA$ is the normalized average of the students' math, ELA, science, and social studies grades, and $TestScores$ is the normalized average of the math and ELA state test scores. When not otherwise stated, we consider that 5\% of students are selected.

The dataset includes demographics, as well as information about the student's current school. We consider the following dimensions of fairness:
\begin{itemize}
    \item {\em Low-income: } in the NYC public school system, $70\%$ of students qualify as low income.
    \item {\em ELL: }students who are English Learners. These students are obviously disadvantaged by an admission method that takes into account ELA (English Language Arts) grades and test scores.
    \item {\em ENI: } the Economic Need Index, a measure of the overall economic need of students attending the same school as the student. ENI is calculated as the percentage of students in the school who have an economic need. A school is defined as high-poverty if it has an Economic Need Index (ENI) of at least $60\%$. 
    \item {\em Special Ed: }students who are receiving special education services.
\end{itemize}

\paragraph{COMPAS}
The COMPAS dataset consists of recidivism data from Broward County Florida as a result of the 2016 ProPublica investigation. The dataset   contains individual demographic information, criminal history, the COMPAS recidivism risk score, arrest records within a 2-year period, for 7214 defendants. COMPAS decile scores, which represent the decile rank of the defendant compared to a target comparison population of defendants, range from 1 to 10. 

We consider the decile score as the ranking function (the lower the better, see discussion in Section~\ref{sec:rescompas}), and compute compensatory bonus points using race as the fairness attribute.

\subsection{Evaluation Parameters}
\label{sec:evaluation_parameters}
\paragraph{Bonus Points.}  The bonus points can be understood as a multiplier over the attribute value. When the attribute is binary, the bonus point value is added to the score of objects with that attribute (e.g., the bonus points for ELL are added to the score of students who are marked as English Learners).  For continuous attributes,  the value of the bonus points is multiplied by the value of the attribute (e.g., the ENI value of the school a student is attending will be multiplied by the ENI bonus points, the resulting product will be added to the score of the student). 

In the final step of the algorithm, we round to the desired bonus point granularity, as decided by stakeholders. For simplicity and efficiency, we restricted bonus points to values with a granularity of 0.5 points in both evaluation scenarios.

\paragraph{DCA vs. Core DCA} In Section~\ref{sec:refine} we evaluate the impact of the refinement step of Algorithm~\ref{refine_function} over the Core DCA algorithm of Algorithm~\ref{algo_SGD}. In the rest of the paper, we use the name DCA to refer to the algorithm {\em with the refinement steps applied}.

\paragraph{Algorithm DCA - Sample Size.} Our rarest fairness category has a frequency of ~10\%, so we picked a sample size of 500 elements to ensure a representation of 50 elements (for our defaults selection percentage of 5\%), enough to show most of the correlation between attributes. 

\paragraph{Algorithm DCA - Learning Rate.} We experimented with different learning rates and settled on 3 sets of DCA with 100 rounds for each learning rate. In the first pass we use a learning rate of 1. This gives us the right general area to search. We use a learning rate of 0.1 to further hone in on the correct location. Then, we take a second pass through the data using a  modern weight updating algorithm, Adam, to find the best bonus point values~\cite{kingma2017adam}. Finally, we take the rolling average of the last 100 points to increase stability and avoid too many random effects of unusual samples near the end.

\subsection{Experimental Environment}
The experiments were all preformed on an Optiplex7060 with 30GB of RAM. The machine has a Intel(R) Core(TM) i7-8700 CPU @ 3.20GHz. Our proposed algorithms were implemented using Python 3.8 and Pandas. The comparison algorithm (Multinomial FA**IR) was implemented in Java using the implementation by the authors of \cite{zehlike2022fair}. 
\section{Evaluation Results}
\label{sec:experiments}

We now discuss detailed evaluation results of DCA using our data sets and over a variety of settings.
\begin{table*}
\vspace{-0.1in}
\label{effectivness-table}
\centering
\begin{tabular}{|p{100pt}|p{50pt}|p{40pt}|p{40pt}|p{40pt}||p{40pt}|}
\hline
{\textbf{Baseline Disparity}}  &Low-Income&ELL&ENI&Special Ed&Norm \\
 \hline
 \hline
 \textbf{Training} 2016-2017 &-0.252&-0.106&-0.176&-0.191&0.377\\
 \hline
 \textbf{Test}  2017-2018&-0.24&-0.105&-0.179&-0.191&0.37\\
 \hline
 \hline

  {\textbf{Core DCA}}  &Low-Income&ELL&ENI&Special Ed&Norm\\
\hline
\textbf{Bonus Points}&2.0&11.0&11.0&14.0&-\\
\hline
\textbf{Training} 2016-2017&0.051&0.018&0.001&0.049&0.073\\
\hline
\textbf{Test} 2017-2018&0.052&-0.006&-0.001&0.029&0.06\\
 
 \hline
 \hline

  {\textbf{DCA}}  &Low-Income&ELL&ENI&Special Ed&Norm\\
\hline
\textbf{Bonus Points}&1.0&11.5&12.0&12.0&-\\
\hline
\textbf{Training} 2016-2017&-0.018&0.001&0.001&-0.014&0.023\\
\hline
\textbf{Test} 2017-2018&-0.01&-0.017&0.005&-0.028&0.034\\

\hline

\end{tabular}
\caption{\em Disparity vectors for the NYC high schools data before and after bonus points}
\label{tab_disparityreduction}
\vspace{-0.15in}
\end{table*}

\subsection{Results on the School Dataset}
\label{sec:resschools}

\subsubsection{Disparity Reduction}

Table~\ref{tab_disparityreduction} shows the disparity values for each target fairness dimension and overall (Norm). The top part of the table shows the Baseline disparity when the school ranking function is used to select $5\%$ of students without any disparity correction. We can see that for both years and for all our fairness targets the disparity is high: for example, low-income students appear $25\%$ less in the selection than in the total population. Overall, the disparity is at 0.37. Note that a disparity of 0 is the goal, the highest the absolute value, the more disparate impacts exist. The sign of the disparity gives the direction of this impact: negative for underrepresentation, positive for overrepresentation.

The bottom portion of the table shows the bonus points produced by the DCA algorithm and the resulting reduced disparity. We can see that for all target attributes the disparity is almost at 0 after the bonus points are applied for both training and test datasets. The overall disparity is also close to 0. The number of bonus points needed to achieve these results is easily understandable: for instance, as ELL students are disadvantaged by the inclusion of ELA grades and scores, they receive 11.5 bonus points on the ranking function to even the playing field. The number of bonus points to address the disparity suffered by low-income students is surprisingly low: giving them just 1 bonus point brings them to statistical parity in the selected set, possibly because the ENI also captures economic disadvantage.

\subsubsection{Utility}
In fair ranking applications, utility measures how much the disparity compensation approach impacts the original rankings. In simpler terms, it measures how far the new ranking is from the uncorrected one.

A common measure of utility is the $nDCG$, or normalized discounted cumulative gain~\cite{jarvelin2002cumulated,zehlike2017fa}. The standard discounted cumulative gain $DCG$ is defined as:
$$ \sum_{i=1}^{k}\frac{w_i}{log_2(i+1)}$$
$nDCG$ is defined as the ratio of the measured $DCG$ to the ideal $DCG$. In fairness ranking applications, the ideal $DCG$ is that of the original ranking before fairness compensation is applied.

\begin{figure*}
\centering
\vspace{-.25in}
\begin{minipage}{.32\textwidth}
  \centering
  \includegraphics[width=\textwidth]{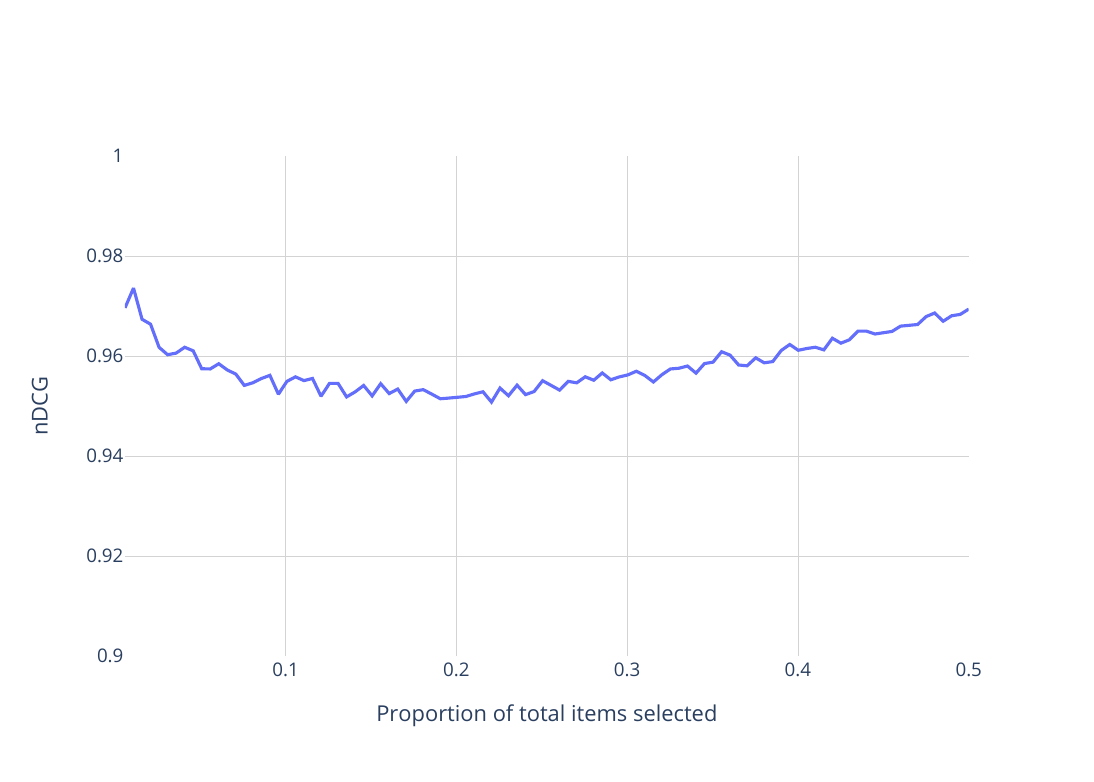}
            \captionof{figure}%
            {{\small $nDCG@k$ on the school data (Test dataset) for varying $k$}}    
            \label{fig:ndcg_k}
\end{minipage}%
\hfill
\begin{minipage}{.32\textwidth}
  \centering
  \includegraphics[width=\textwidth]{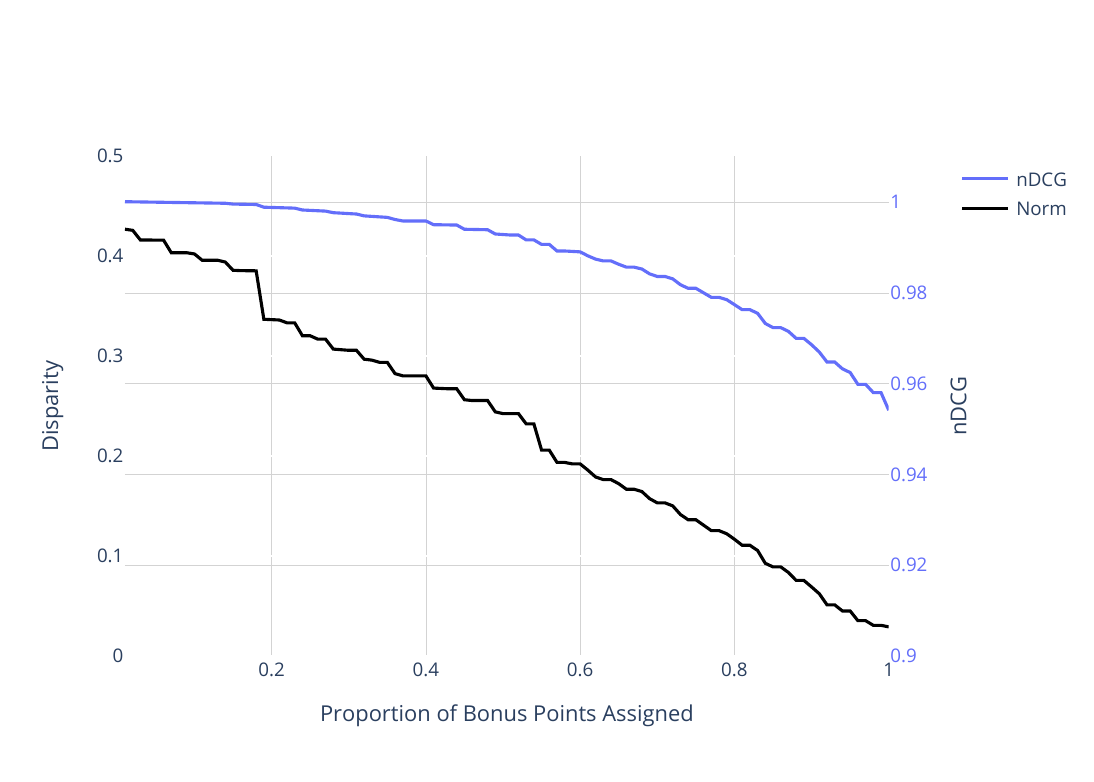}
            \captionof{figure}%
            {{\small $nDCG@k$ and disparity norm on the school data (Test dataset) for varying proportions of total recommended bonus points}}    
            \label{fig:utility_adjust}
\end{minipage}%
\hfill
\begin{minipage}{.32\textwidth}
  \centering
  \includegraphics[width=\textwidth]{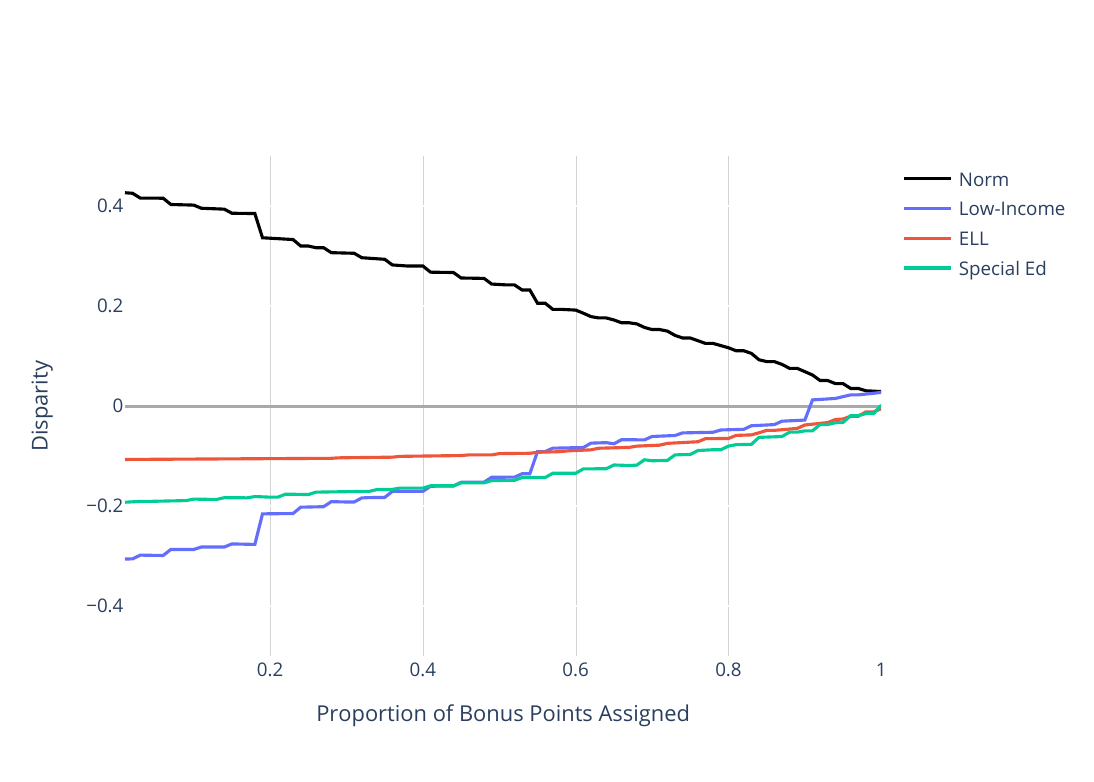}
            \captionof{figure}%
            {{\small Disparity on the school data (Test dataset) for varying proportions of total recommended bonus points }}    
            \label{fig:full_disp_adjust}
\end{minipage}
\end{figure*}

 A score of 1 would mean that the fairness compensation comes at no loss in utility at all: the ranking is unchanged. For our default selection set of $5\%$ of students, the $nDCG@0.05$ of DCA is  0.957. This is comparable to other fair ranking  algorithms that handle multiple  fairness dimensions.  \cite{zehlike2022fair}. Figure~\ref{fig:ndcg_k} reports the  $nDCG@k$ where $k$ represents the proportion of selected students, showing good utility for all selection percentages.
    
DCA can  easily be calibrated for different desired fairness thresholds or utility values. Bonus points may be adjusted by a weight multiplicative factor to reduce the importance of the bonus points and increase the utility (as measured by $nDCG$). The correct proportion of bonus points to apply can be selected through a binary search. Figure \ref{fig:utility_adjust} shows the impact on  utility and disparity of applying a reducing weight to bonus points. 

Figure \ref{fig:full_disp_adjust} shows a detailed breakdown of the disparity as we adjust the total proportion of the bonus points. The step nature of the function is due to our restriction that the bonus points be a multiple of 0.5. The function is near linear, applying half of the optimal disparity reduction bonus points, provides about half the disparity reduction. This shows that DCA can be easily adjusted to  provide a solution for any given utility or fairness threshold.

\subsubsection{Effect of Varying the Percentage of Selected Objects}
\label{sec:varying}
    \begin{figure*}
        \centering
        \vspace{-0.24in}
        \begin{subfigure}[t]{0.32\textwidth}
            \centering
            \includegraphics[width=\textwidth]{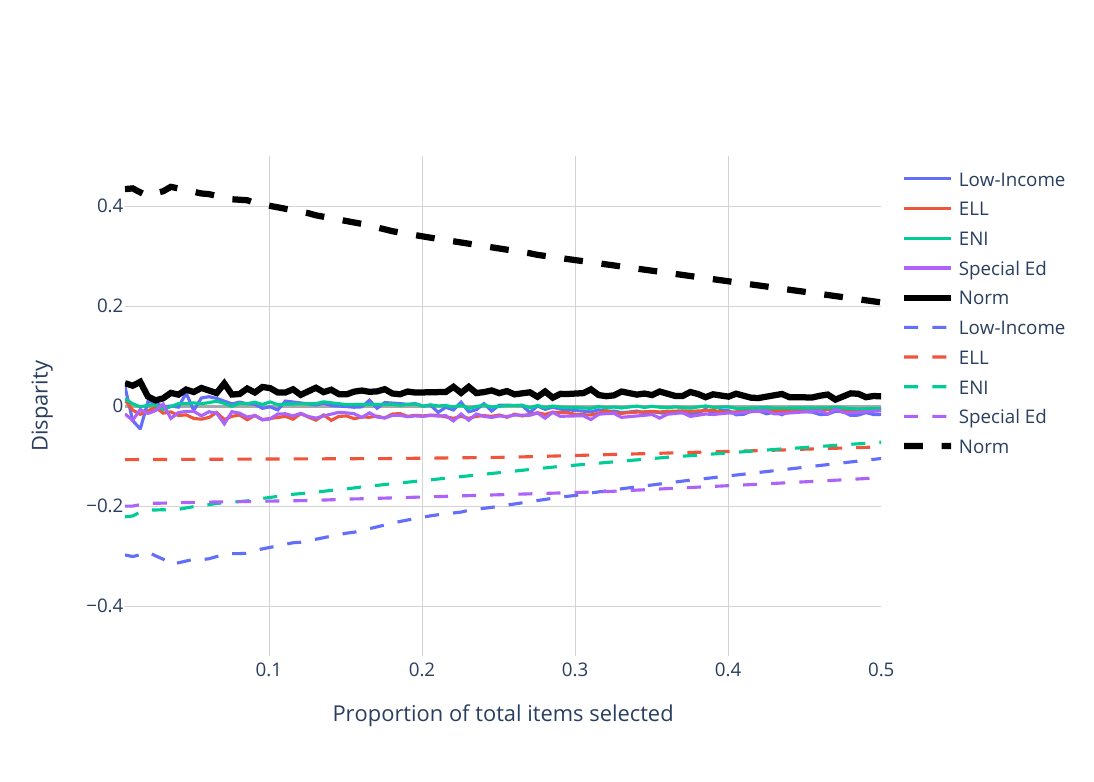}
            \caption[]%
            {{\small Disparity adjusted for k on the school data when k is known in advance}}    
            \label{fig:school_all_k}
        \end{subfigure}
        \hfill
        \begin{subfigure}[t]{0.32\textwidth}   
            \centering 
            \includegraphics[width=\textwidth]{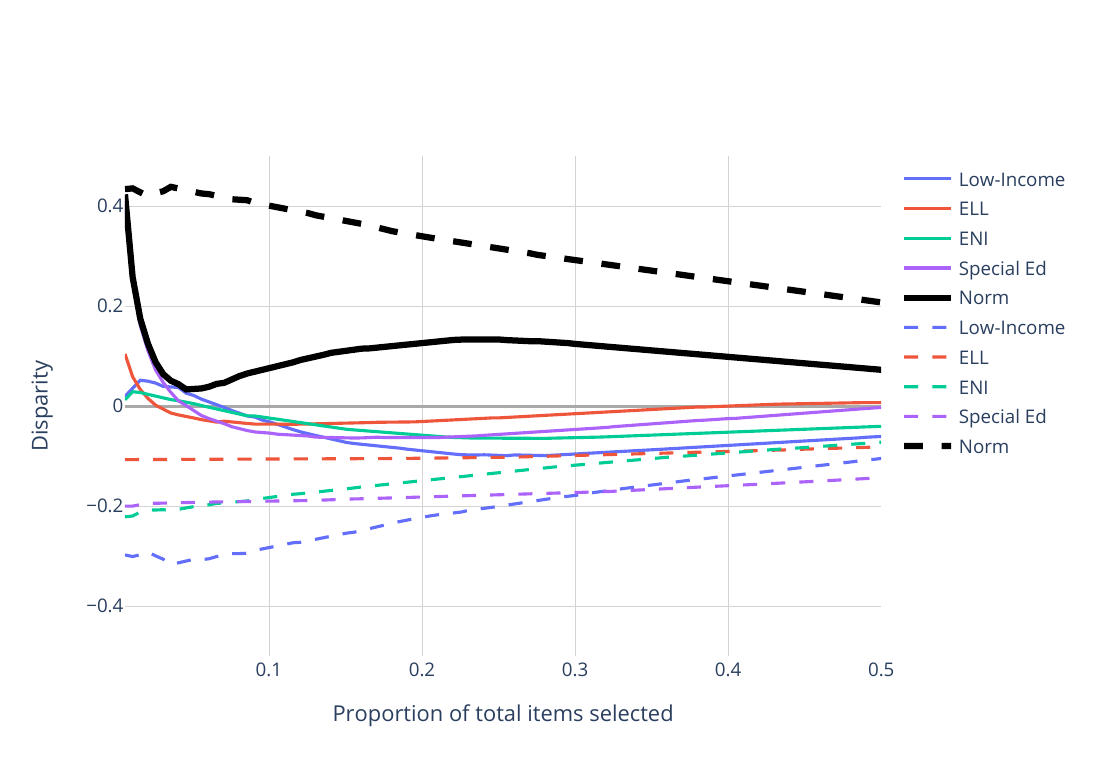}
            \caption[]%
            {{\small The disparity across all k when k is assumed to be 5\%} The bonus vector for this graph is: \{Special-Ed:13, Low-Income:1.5, ELL:10.5, ENI:12.0\}}    
            \label{fig:school_5}
        \end{subfigure}
        \begin{subfigure}[t]{0.32\textwidth}   
            \centering 
            \includegraphics[width=\textwidth]{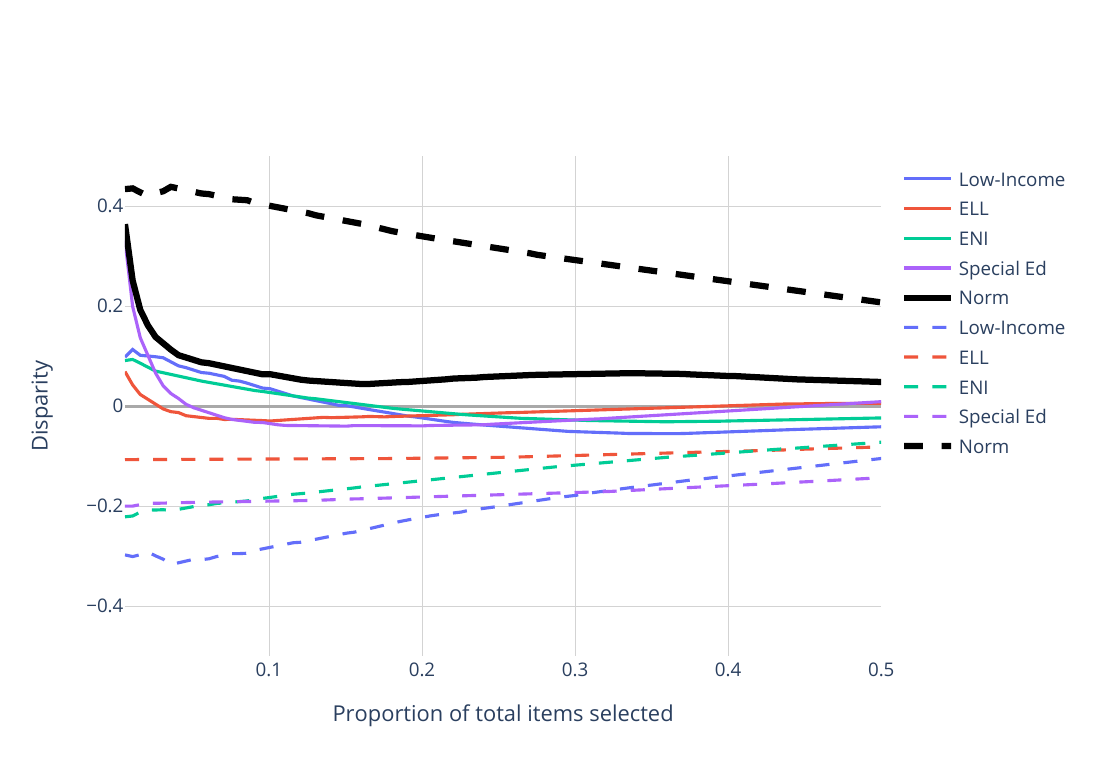}
            \caption[]%
            {{\small The disparity across all k when k is assumed to be logarithmically discounted for k$<$0.5 the bonus vector, In this case, the bonus vector is \{Special-Ed:13.5, Low-Income:2, ELL:9.5, ENI:18.5\}}}    
            \label{fig:log_discount k}
        \end{subfigure}
        \caption[ ]
        {\small Experiments using DCA on the School dataset } 
        \label{fig:School_DCA}
    \end{figure*}

Some applications determine the number of selected objects in advance (e.g, vaccine allocation). Others may be impacted that external factors that will vary the number of selected objects, for example through a waitlist process. As explained in Section~\ref{sec:motivations}, the NYC school admissions are handled by a matching algorithm. Schools do not know in advance how far down their ranked list they will accept students as this will depend on many factors: student choices, other schools' rankings and enrollment targets.

As discussed in Section~\ref{sec:log_discount}, our algorithm can account for variations in the percentage of selected objects $k$ in different ways: 
\begin{itemize}
    \item If $k$ is known, our algorithm can optimize for the specific value and give excellent results. In Figure \ref{fig:school_all_k}, the disparity before (dashed line) and after (full line) correction is shown for varying $k$. In every case, given the selectivity, DCA succeeds in essentially eliminating disparity. \remove{The associated bonus points for each value of $k$ are shown in Figure \ref{fig:school_bonus}.}
    \item If $k$ is not known \yehuda{at the time of bonus point assignment} but can be approximated, DCA can be optimized for the approximation and results in good results \yehuda{when this estimate is close to the real value} (Figure \ref{fig:school_5}). The disparity results degrade however when $k$ is not estimated properly.
    \item If $k$ is unknown, or several different $k$ values are important, we use our logarithmically-discounted  approach (Section~\ref{sec:log_discount})  to set bonus points to the setting that will provide the best disparity outcome for a weighted average of many different $k$ values in the ranked list. This means that DCA's goal is to minimize the weighted average of disparities across many values of $k$ instead of only minimizing the disparity at a specific $k$. However, if the exact value of k is known \yehuda{when bonus points are chosen}, selecting a bonus vector that minimizes the disparity  for that exact value of $k$ provides better results for that specific $k$, at the cost of a higher average across the other values of $k$. This can be seen by comparing the disparity in Figure \ref{fig:school_5} and Figure \ref{fig:log_discount k} at $k=0.05$. While Figure \ref{fig:log_discount k} has a lower disparity at most $k$, Figure \ref{fig:school_5} shows DCA specifically targeting $k$ near 0.05 and has better results when $k$ is near 0.05.
\end{itemize}

   \begin{figure*}
   \vspace{-.24in}
    \begin{minipage}{.32\textwidth}
  \centering
  \includegraphics[width=\textwidth]{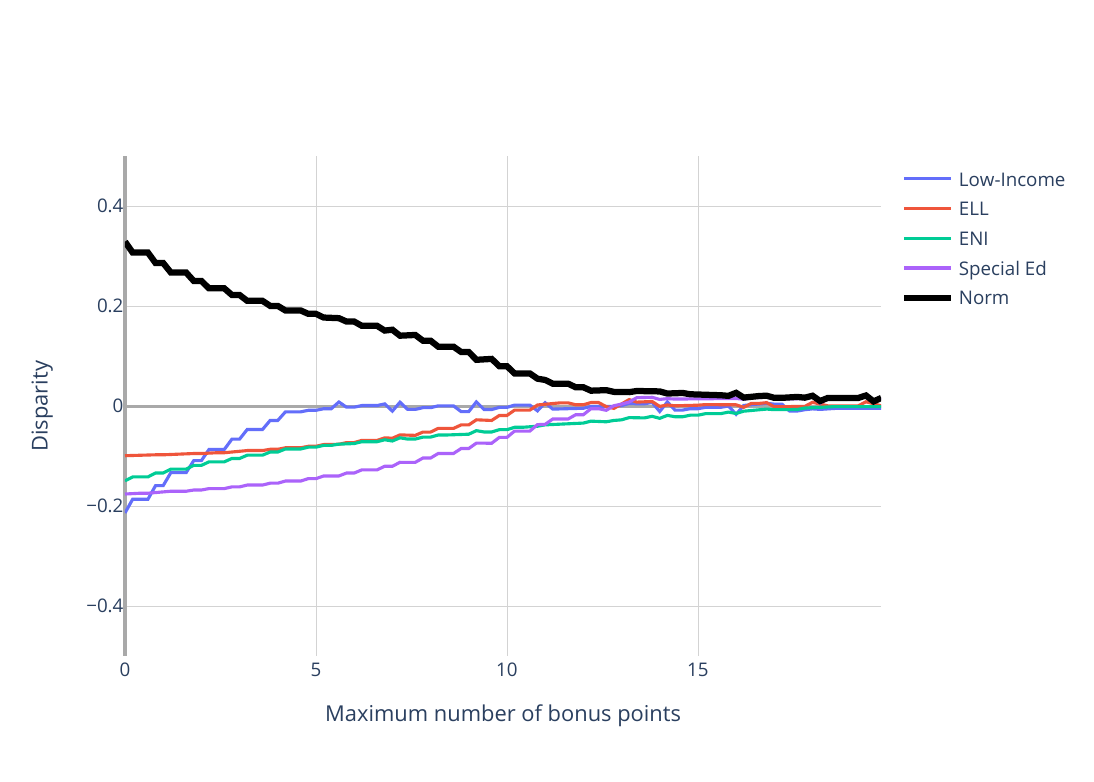}
    \captionof{figure}%
    {{\small Log-Discounted disparity when there is a maximum number of bonus points}}    
    \label{fig:max_bonus}

\end{minipage}
\hfill
   \begin{minipage}{.32\textwidth}
  \centering
  \includegraphics[width=\textwidth, trim={0 1cm 0 0},clip]{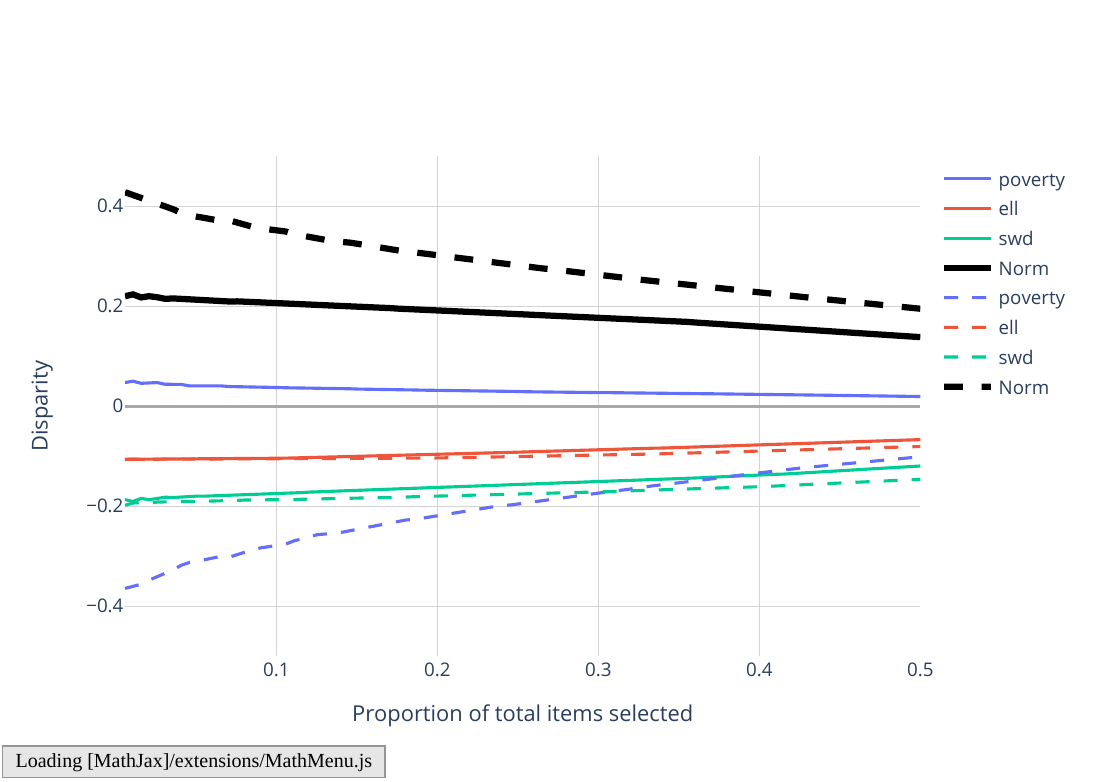}
        \captionof{figure}%
        {{\small \yehuda{The disparity reduction achieved by a simple quota system } }}    
        \label{fig:school_quota}

    \end{minipage}
    \hfill
    \begin{minipage}{.32\textwidth}
  \centering
  \includegraphics[width=\textwidth]{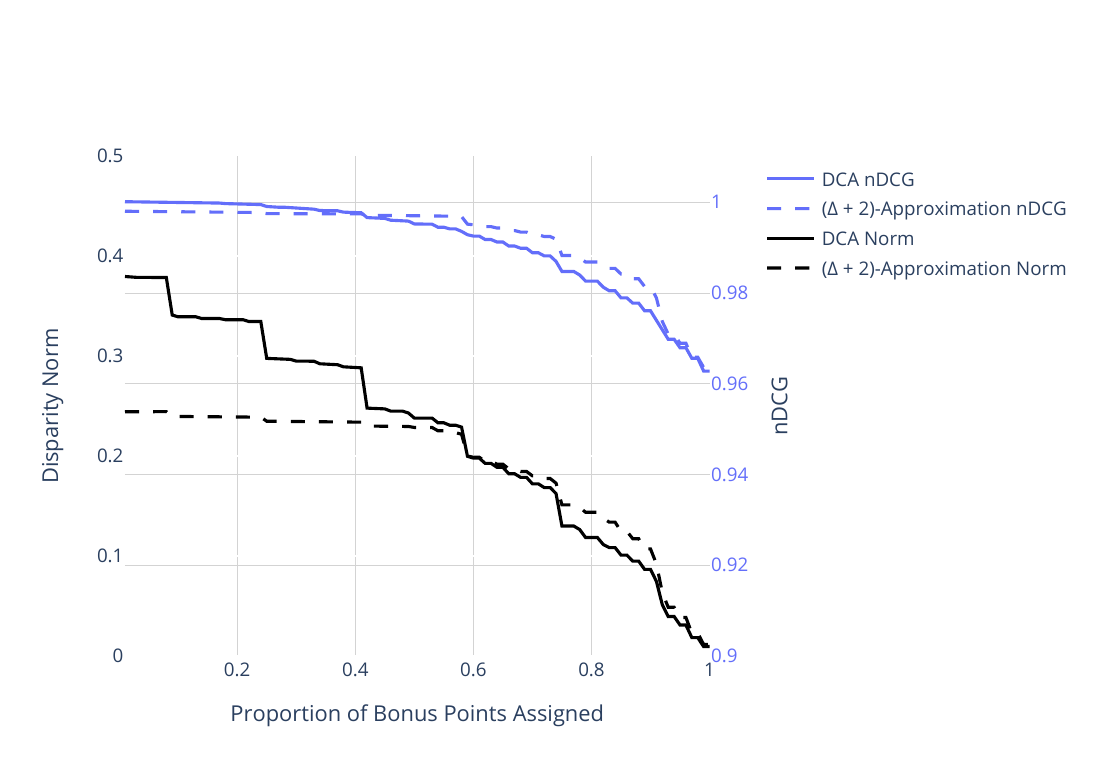}\vspace{0.1in}
            \captionof{figure}%
            {{\small Accuracy vs. disparity for both DCA and the ($\Delta$  + 2)-approximation\  algorithm (training Dataset)}}    
            \label{fig:utility_comparison}
\end{minipage}

\end{figure*}

\subsubsection{Maximum Bonus Limits}

DCA can easily be adapted to bound the bonus values it allocates using preset minimum and maximum bonus values. All experiments in this paper cap DCA to never give negative bonuses, as these can be perceived as penalties. If desired, maximum bonuses can also be set. The number of bonus points can be capped at every refinement step; this may cause adjustments in correlated non-capped attributes. Figure \ref{fig:max_bonus} shows the logarithmically discounted disparity for $k<0.05$ when the bonus amount is limited between 0 and 20. The resulting disparity is obviously impacted, with worst results when the maximum number of bonus points is small, however  as the maximum number of bonus points increases, the disparity gets smaller.

\subsubsection{Impact of the Refinement Step}
\label{sec:refine}
    \begin{figure*}
    \vspace{-.23in}
    \begin{minipage}{.67\textwidth}
        \centering
        \begin{subfigure}[b]{0.48\textwidth}
            \centering
            \includegraphics[width=\textwidth]{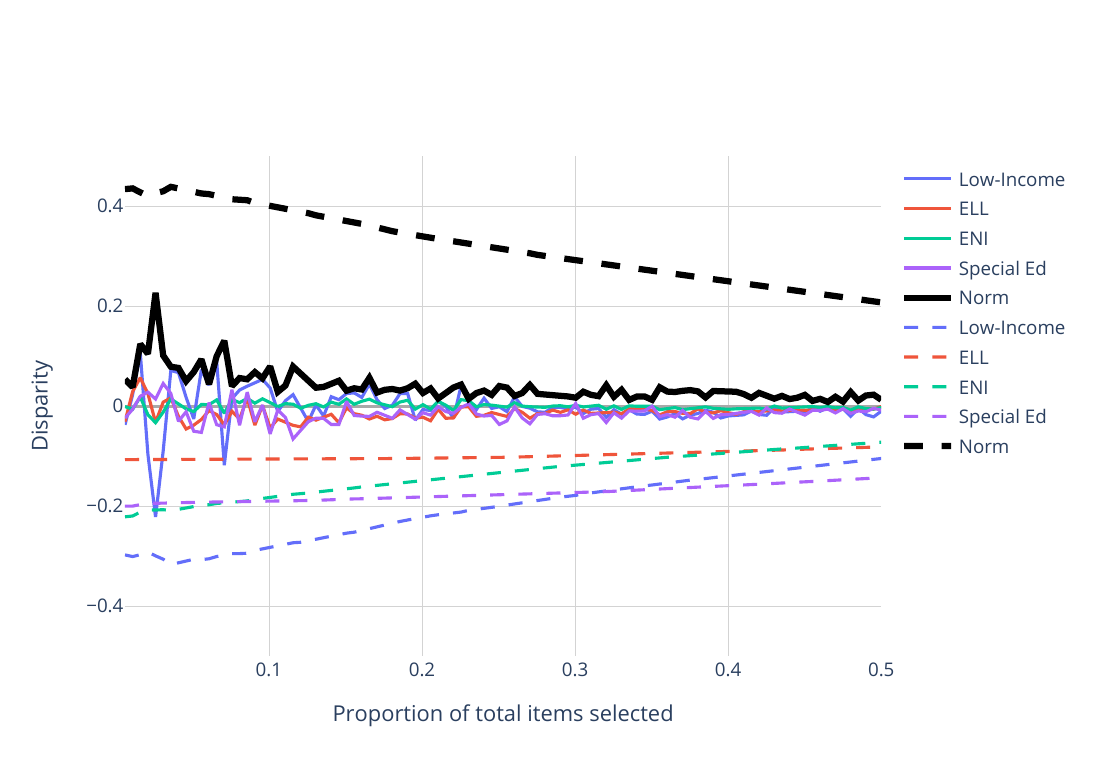}
            \caption[]%
            {{\small }}    
            \label{fig:school_all_k_noref}
        \end{subfigure}
        \hfill
        \begin{subfigure}[b]{0.48\textwidth}  
            \centering 
            \includegraphics[width=\textwidth]{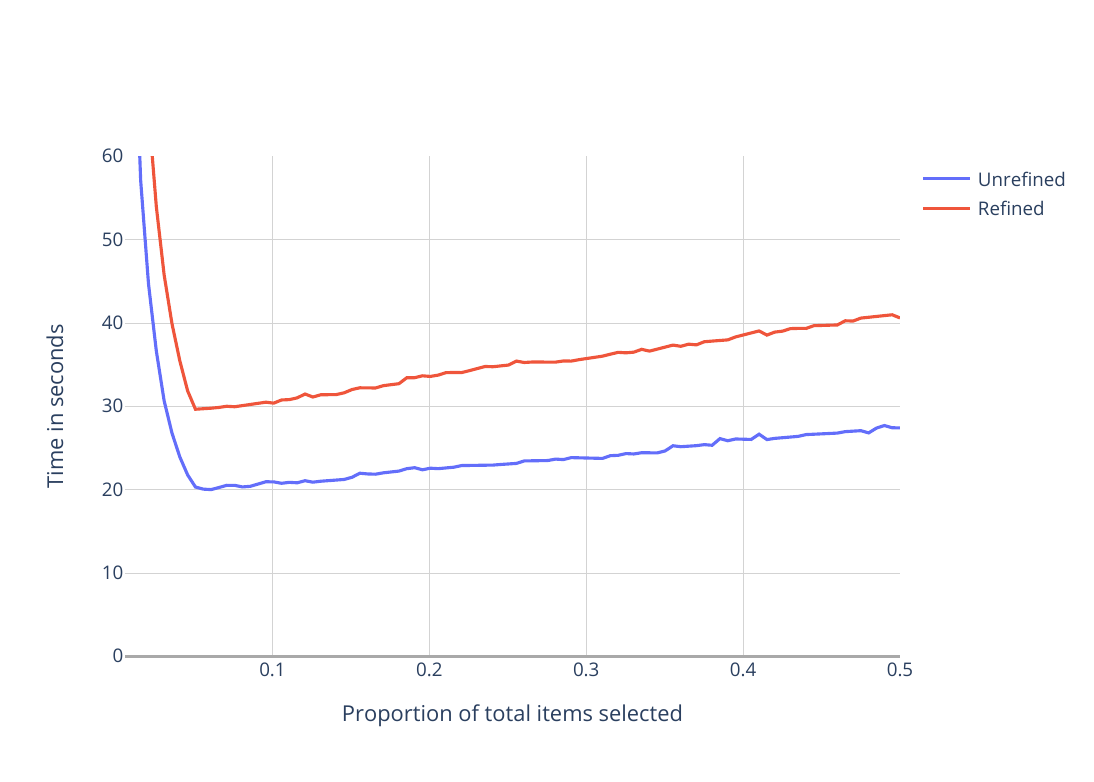}
            \caption[]%
            {{\small }}    
            \label{fig:school_time_comparison}
        \end{subfigure}
        \vskip\baselineskip
        \vspace{-0.2in}
        \caption[ ]
        {\small Evaluation of the DCA Refinement Step on the School dataset. (a) Disparity adjusted for k on the school data when k is known in advance, using Core DCA without refinement, (b) Time taken for each k in both the unrefined (Core DCA) and refined version (DCA).} 
        \label{fig:no_refine}
        \end{minipage}
        \hfill
\begin{minipage}{.32\textwidth}
  \centering
  \includegraphics[width=\textwidth]{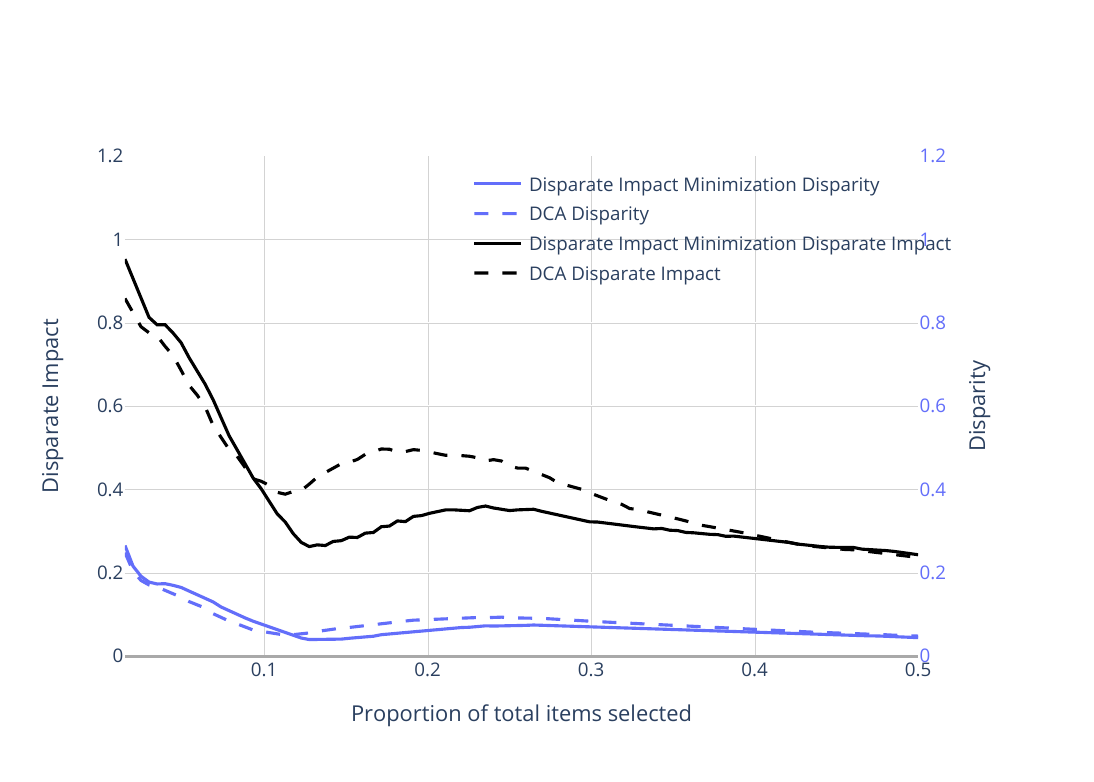}
    \captionof{figure}%
   {{\small Norm after optimization with DCA for Disparity (dotted lines), or  Disparate Impact (solid lines). Disparity in blue, Disparate Impacts in black.}}    
    \label{fig:Disparate_impact}
    \end{minipage}%
    \end{figure*}

Table \ref{tab_disparityreduction} shows that the refinement step of Algorithm~\ref{refine_function} results in improvements over the results of the Core DCA Algorithm~\ref{algo_SGD} alone. Over the school dataset, those improvements are about threefold. Figure \ref{fig:school_all_k_noref} shows the same settings as Figure~\ref{fig:school_all_k}, but without the refinement step. We see that in addition to better Disparity compensation, the refinement step produces smoother results.

Figure \ref{fig:school_time_comparison} shows the time taken by both the Core DCA and DCA with refinement. In most cases, the refinement step takes approximately 10 seconds. As seen in the figure, the cost is higher for small values of $k$. This is due to the fact that the sample size needed for DCA depends on $max(\frac{1}{k},\frac{1}{r})$. When k is small the sample size therefore needs to be increased, which leads to longer execution times. Once $\frac{1}{k}$ becomes small enough at 5\%, the sample size is based on $r$, the frequency of the least
common group in the dataset, which is the same for all settings over the dataset. As k increases however, the computation of the centroid as part of the Disparity $\vec{D_k}$ computation takes longer as more elements are considered.

These results show the benefit of the refinement step.
In cases where faster execution times are desireable, this step can be omitted with some loss in quality.

\subsection{Results on the COMPAS dataset}
\label{sec:rescompas}

    \begin{figure*}
    \vspace{-.23in}
        \centering
        \begin{subfigure}[b]{0.32\textwidth}
            \centering
            \includegraphics[width=\textwidth]{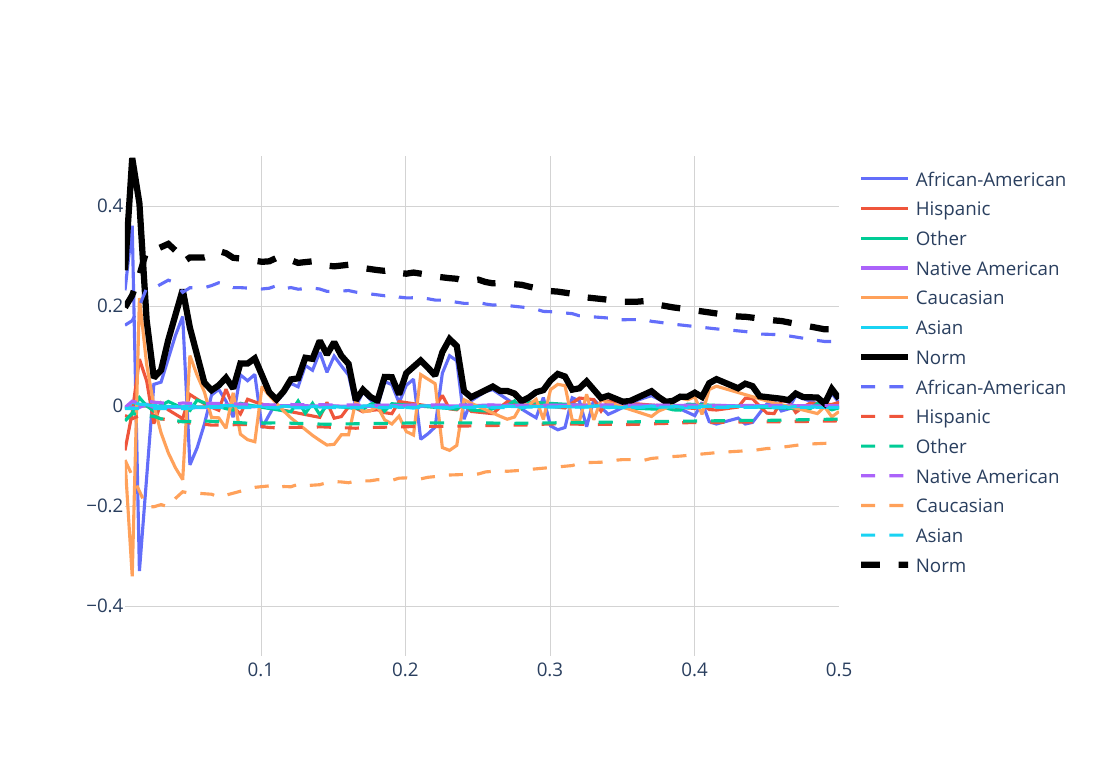}
            \caption[]%
            {{\small The Disparity of the COMPAS algorithm when bonus points are added to the decile scores and computed for each $k$}}    
            \label{fig:compas_unchanges}
        \end{subfigure}
        \hfill
        \begin{subfigure}[b]{0.32\textwidth}  
            \centering 
            \includegraphics[width=\textwidth]{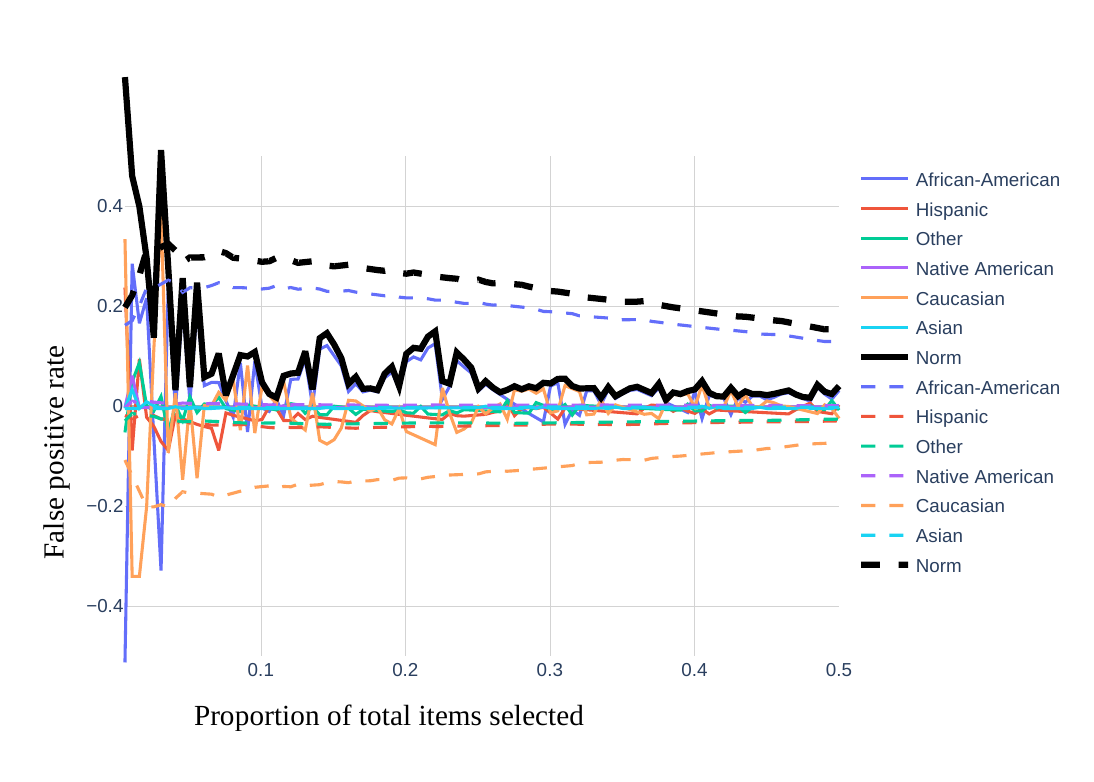}
            \caption[]%
            {{\small \yehuda{The False Positive Rate of the COMPAS algorithm when bonus points are added to the decile scores and computed for each  $k$} }}    
            \label{fig:compas_FPR}
        \end{subfigure}
        \hfill
                \begin{subfigure}[b]{0.32\textwidth}  
            \centering 
            \includegraphics[width=\textwidth]{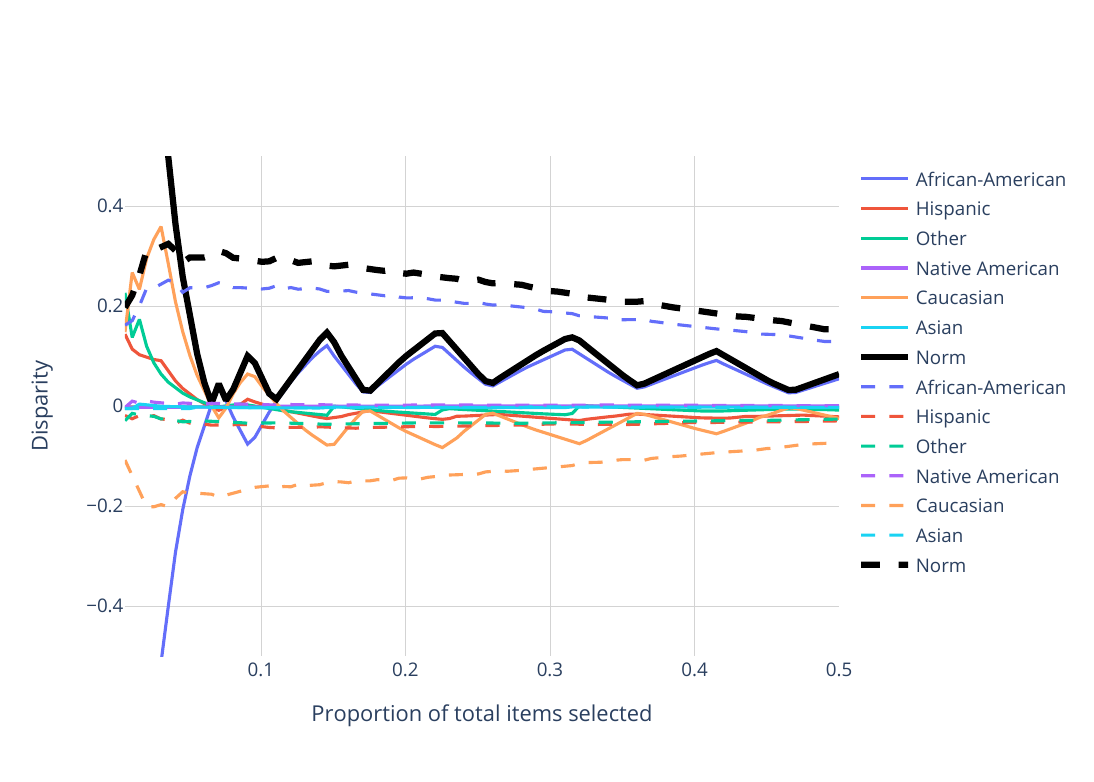}
            \caption[]%
            {{\small The Disparity of the COMPAS algorithm when bonus points are computed once using the log discount mode of DCA}}    
            \label{fig:compas_log_discount}
        \end{subfigure}
                \caption[ ]
        {\small Experiments using DCA on the COMPAS dataset } 
        \label{fig:COMPAS_DCA}

\end{figure*}

Figure~\ref{fig:compas_unchanges} shows the baseline disparity of the COMPAS decile scores based on race (dashed line). The disparity is notable for Black people who are significantly more likely to be flagged for recidivism risk, and for white people who are significantly less likely. The bonus point compensation offered by DCA (full line) allows to significantly reduce that disparity.

A difficulty with trying to address disparity in the COMPAS data is that its scores are very coarse: they are only 10 possible scores, which makes it difficult to have an impact with bonus points. This can be seen most clearly when the log discount mode of DCA is used in Figure \ref{fig:compas_log_discount}: As each new bucket moves into the selected set, the disparity moves sharply. \yehuda{However, shown in figures \ref{fig:compas_unchanges} and \ref{fig:compas_log_discount}, all cases still result in significant disparity reduction.}

\remove{
There are a few ways to mitigate this problem. First, we note that the decile score is not the actual score computed by COMPAS, but a decile-rank representation of that score. If the underlying score, or its distribution,  were available, we could simply convert the decile into its underlying score. We simulated this approach by assuming that COMPAS scores were taken from a normal distribution and then selecting from each decile a score from the underlying distribution (using uniform  random sampling). DCA was applied on these scores, the results are shown in Figure~\ref{fig:compas_underlying}. Another option is to accept reduced disparity mitigation. Since the underlying COMPAS algorithm has large buckets, it makes sense for the fairness compensation. As is shown in figures \ref{fig:compas_unchanges} and \ref{fig:compas_log_discount}  this still results in significant disparity reduction.
}

\subsection{Comparison with Existing Approaches and Metrics}
\label{sec:Comparison}
\subsubsection{Comparison to a Real-world Quota System}

\yehuda{As discussed earlier, quota-based
approaches are difficult to extend to the multinomial case. Many real-world settings, such as the NYC school system, use one single quota for all the different fairness dimensions. Figure \ref{fig:school_quota} shows that this approach does result in  disparity reduction, but does not achieve results as good as DCA (Figure~\ref{fig:School_DCA}).}  
\subsubsection{Comparison to Multinomial FA**IR}
We also compared DCA with the Multinomial FA**IR method~\cite{zehlike2022fair} on the school dataset. Multinomial FA**IR is a post-processing method that re-ranks the dataset with fairness guarantees. Multinomial FA**IR only works on non-overlapping fairness parameters, so we looked at the Cartesian product of all our parameters and picked the 3 most-discriminated against subgroups as our barometers of fairness, as suggested in~\cite{zehlike2022fair}. Because of efficiency limitations of Multinomial FA**IR, we were only able to run their code on a single district of NYC schools, consisting of 2,500 students, instead of the whole dataset. We report our results in Table~\ref{tab:mfair}. We see that, while both methods significantly improve disparity, DCA results in better outcomes due to its ability to handle overlapping subgroups.

\begin{table}
\centering
\begin{tabular}{|p{60pt}|p{32pt}|p{30pt}|p{30pt}||p{30pt}|}
\hline
{}  &Low-Inc&ELL&Sp. Ed&Norm \\
 \hline 
 \textbf{Baseline } &-0.262&-0.036&-0.179&0.320\\

\hline \hline
\textbf{Bonus Points}&2&9&5&-\\
\hline 
\textbf{DCA}&0.009&-0.011&0.001&0.007\\

\hline
\hline
\textbf{Mult. FA**IR}&-0.084&-0.036&-0.052&0.105\\
\hline
\end{tabular}
\caption{\em Comparison of DCA and Multinomial FA**IR}
\label{tab:mfair}
\vspace{-0.2in}

\end{table}

\subsubsection{Comparison to ($\Delta$ + 2)-approximation\  algorithm}
\label{sec:deltaplustwo}
Because of the time complexity of Multinomial FA**IR, we also include a comparison to a faster approximation algorithm, ($\Delta$  + 2), from \cite{celis2017ranking}. This algorithm works by looking at all (position,item) pairs and greedily selecting the one that most improves the utility (in our case measured by nDCG) without violating a preset (input) fairness constraints on the maximum number of items of each type. To make a fair comparison, we gave ($\Delta$ + 2) the disparity achieved by DCA as its input preset fairness constraint. Unlike DCA, ($\Delta$ + 2) is a post-processing step which only works on existing results; therefore we compared it to the results of DCA on a single year. 
As is shown in Figure \ref{fig:utility_comparison}, ($\Delta$ + 2) achieves results very similar  to DCA. In terms of efficiency, this algorithm depends heavily on the proportion of items selected $k$. For small values of $k$, such as 5\%, it performs similarly to DCA, around 30 seconds, for larger values, such as 30\%, it takes around 30 minutes, making DCA a faster option. 

\remove{In general, our results  show that across a number of settings, in utility, fairness and efficiency, DCA performs similarly or better than similar non-explainable counterparts. In addition, DCA allows for {\em transparent and explainable fairness mitigation}.}

\subsubsection{Exposure}
\label{sec:exposure}

Exposure is a common metric for measuring fairness in ranking. It is defined as the sum of the probability of an object having a position in the ranked order times the value of that position. The value of a position has been defined in different ways in different sources; we used the definition from \cite{gupta2021online}.  They define exposure as $$ Exposure(G|R) = \sum_{i\in G} \frac{1}{lg(r(i) + 1)}$$ Where G is a group, R is a ranking, and r(i) is the rank of an object. They define a fairness metric based on this definition: demographic disparity constraint or DDP, defined as 
$$DDP(R) = max(G_j,G_k) \frac{exposure(G_j|R)}{|G_j|} - \frac{exposure(G_k|R)}{|G_k|} $$
Intuitively, this means that no group should have very different exposure from any other group. A value of zero would mean perfect fairness. The exact values are not comparable across datasets of different sizes since the value of exposure shrinks as the dataset grows. 

We calculated the exposure value on the school dataset without the ENI attribute, as DDP does not handle non-binary fairness attributes. Since exposure considers the entire ranking, the logarithmic discounting mode of DCA was used. The resulting bonus point vector was \texttt{\{'Special-Ed': 14, 'Low-Income': 5, 'ELL': 11\}}. 

Under the baseline disparity setting, the DDP value is 0.00899. After DCA compensatory bonus points are applied, the DDP becomes 0.00166. This 5-fold reduction in DDP is in line with the disparity experiments from Section~\ref{sec:resschools}, confirming that important improvements in fairness can be achieved with reasonable size bonus point vectors.
\subsubsection{Using DCA with other fairness metrics}

Our DCA algorithm can be used to minimize fairness metrics other than disparity. A limitation is that the minimization metric must be represented as the norm of a vector, and it must provide bounds between -1,1 with -1 representing complete unfairness to one group and 1 representing complete unfairness to another, and 0 representing fairness. 

To show the behavior of DCA with other metrics, we have implemented a slight variation of one of the most popular fairness metrics: {\em disparate impact (DI)}. DI sets limits on the ratio of positive (selected) objects in the protected and unprotected groups. We use the slightly modified version from\cite{zafar2017fairness}. Specifically, for each fairness dimension, the disparate impact is measured as:
$$
min \left( \frac{P(O=1|F=0)}{P(O=1|F=1)}, \frac{P(O=1|F=1)}{P(O=1|F=0)} \right)
$$
Where F=0 represents the object not having a protected (fairness) attribute (e.g., not being low-income) and O=1 represents the object being selected. To make it usable by  DCA, we had to scale it to the [-1;1] range. With this modification, disparate impact can be directly applied in the discrete case (including the logarithmically-discounted variation) leading to a bonus vector of {'Special Ed': 14 pts, 'Low-Income': 5.5 pts, 'ELL': 12.5 pts} compared to the similar bonus vector using disparity of  {'Special Ed': 14 pts, 'Low-Income': 5 pts, 'ELL': 11.5 pts}. We show the comparison of using Disparity and DI with DCA Figure~\ref{fig:Disparate_impact}. Both versions perform similarly. The disparate impact version of DCA took 164 seconds compared to  64 seconds for regular DCA with these settings.

\yehuda{This type of expansion is not limited to statistical parity measures: When data is available, DCA can also be used with equalized odds measures such as the False Positive Rate. The FPR is defined as the proportion of real negative cases that were misidentified as positive by the algorithm. Disparities in this rate between different groups is one of the original criticisms of the COMPAS algorithm. To minimize this difference we subtract the overall FPR from the per-group FPR. This difference has the required properties for DCA. \remove{It is between -1,1, since the FPR is between 0 and 1. -1 represents that all false positives are in other groups, 1 represents that all false positives are in this group, and 0 representing an equal FPR for this group and the distribution as a whole.} Using this metric with DCA results in an algorithm that finds the number of bonus points to minimize the differences in FPRs between groups. Figure \ref{fig:compas_FPR}, shows that the FPR difference is reduced across a range of $k$s.}

\section{Conclusion}
\label{sec:conclusion}

We presented DCA, an algorithm to address disparity in outcomes of ranking processes using compensatory bonus points. We showed that DCA, by relying on a sampling-based approach, successfully reduces disparity in a wide range of settings, while being significantly more efficient than state-of-the-art approaches, running in sub-linear time. This makes DCA a good candidate for iterative processes that would allow users to identify the ranking function that best fits their needs while checking for its fairness impacts and the required compensatory bonus points.

Our approach relies on the use of compensatory bonus points, a departure from previous work, which has mostly focused on modifying the ranking function directly, or on the use of quotas. A significant advantage of compensatory bonus points is that they are transparent, interpretable, and easily explainable to all stakeholders.

\bibliographystyle{IEEEtran}
\bibliography{bonus, Ranking}

\begin{thebibliography}{10}
\providecommand{\url}[1]{#1}
\csname url@samestyle\endcsname
\providecommand{\newblock}{\relax}
\providecommand{\bibinfo}[2]{#2}
\providecommand{\BIBentrySTDinterwordspacing}{\spaceskip=0pt\relax}
\providecommand{\BIBentryALTinterwordstretchfactor}{4}
\providecommand{\BIBentryALTinterwordspacing}{\spaceskip=\fontdimen2\font plus
\BIBentryALTinterwordstretchfactor\fontdimen3\font minus
  \fontdimen4\font\relax}
\providecommand{\BIBforeignlanguage}[2]{{%
\expandafter\ifx\csname l@#1\endcsname\relax
\typeout{** WARNING: IEEEtran.bst: No hyphenation pattern has been}%
\typeout{** loaded for the language `#1'. Using the pattern for}%
\typeout{** the default language instead.}%
\else
\language=\csname l@#1\endcsname
\fi
#2}}
\providecommand{\BIBdecl}{\relax}
\BIBdecl

\bibitem{sonmez2019affirmative}
T.~S{\"o}nmez, M.~B. Yenmez \emph{et~al.}, \emph{Affirmative action with
  overlapping reserves}.\hskip 1em plus 0.5em minus 0.4em\relax Boston College,
  2019.

\bibitem{Gale2020ExplainingMR}
A.~Gale and A.~Marian, ``Explaining monotonic ranking functions,''
  \emph{Proceedings of the VLDB Endowment}, vol.~14, no.~4, pp. 640--652, 2020.

\bibitem{yang2017measuring}
K.~Yang and J.~Stoyanovich, ``Measuring fairness in ranked outputs,'' in
  \emph{Proceedings of the 29th international conference on scientific and
  statistical database management}, 2017, pp. 1--6.

\bibitem{zehlike2021fairness}
M.~Zehlike, K.~Yang, and J.~Stoyanovich, ``Fairness in ranking: A survey,''
  2021.

\bibitem{Fair_Classification}
\BIBentryALTinterwordspacing
M.~T. Islam, A.~Fariha, A.~Meliou, and B.~Salimi, ``Through the data management
  lens: Experimental analysis and evaluation of fair classification,'' in
  \emph{Proceedings of the 2022 International Conference on Management of
  Data}, ser. SIGMOD '22.\hskip 1em plus 0.5em minus 0.4em\relax New York, NY,
  USA: Association for Computing Machinery, 2022, p. 232–246. [Online].
  Available: \url{https://doi.org/10.1145/3514221.3517841}
\BIBentrySTDinterwordspacing

\bibitem{pitoura2021fairness}
E.~Pitoura, K.~Stefanidis, and G.~Koutrika, ``Fairness in rankings and
  recommenders: Models, methods and research directions,'' in \emph{2021 IEEE
  37th International Conference on Data Engineering (ICDE)}, 2021, pp.
  2358--2361.

\bibitem{lahoti2019ifair}
P.~Lahoti, K.~P. Gummadi, and G.~Weikum, ``ifair: Learning individually fair
  data representations for algorithmic decision making,'' in \emph{2019 ieee
  35th international conference on data engineering (icde)}.\hskip 1em plus
  0.5em minus 0.4em\relax IEEE, 2019, pp. 1334--1345.

\bibitem{kamiran2012data}
F.~Kamiran and T.~Calders, ``Data preprocessing techniques for classification
  without discrimination,'' \emph{Knowledge and information systems}, vol.~33,
  no.~1, pp. 1--33, 2012.

\bibitem{feldman2015certifying}
M.~Feldman, S.~A. Friedler, J.~Moeller, C.~Scheidegger, and
  S.~Venkatasubramanian, ``Certifying and removing disparate impact,'' in
  \emph{proceedings of the 21th ACM SIGKDD international conference on
  knowledge discovery and data mining}, 2015, pp. 259--268.

\bibitem{calmon2017optimized}
F.~Calmon, D.~Wei, B.~Vinzamuri, K.~Natesan~Ramamurthy, and K.~R. Varshney,
  ``Optimized pre-processing for discrimination prevention,'' \emph{Advances in
  neural information processing systems}, vol.~30, 2017.

\bibitem{salimi2019interventional}
B.~Salimi, L.~Rodriguez, B.~Howe, and D.~Suciu, ``Interventional fairness:
  Causal database repair for algorithmic fairness,'' in \emph{Proceedings of
  the 2019 International Conference on Management of Data}, 2019, pp. 793--810.

\bibitem{radlinski2008learning}
F.~Radlinski, R.~Kleinberg, and T.~Joachims, ``Learning diverse rankings with
  multi-armed bandits,'' in \emph{Proceedings of the 25th international
  conference on Machine learning}.\hskip 1em plus 0.5em minus 0.4em\relax ACM,
  2008, pp. 784--791.

\bibitem{asudeh2019designing}
A.~Asudeh, H.~Jagadish, J.~Stoyanovich, and G.~Das, ``Designing fair ranking
  schemes,'' in \emph{Proceedings of the 2019 International Conference on
  Management of Data}, 2019, pp. 1259--1276.

\bibitem{celis2017ranking}
L.~E. Celis, D.~Straszak, and N.~K. Vishnoi, ``Ranking with fairness
  constraints,'' \emph{arXiv preprint arXiv:1704.06840}, 2017.

\bibitem{zehlike2017fa}
M.~Zehlike, F.~Bonchi, C.~Castillo, S.~Hajian, M.~Megahed, and R.~Baeza-Yates,
  ``Fa* ir: A fair top-k ranking algorithm,'' in \emph{Proceedings of the 2017
  ACM on Conference on Information and Knowledge Management}, 2017, pp.
  1569--1578.

\bibitem{zehlike2022fair}
M.~Zehlike, T.~S{\"u}hr, R.~Baeza-Yates, F.~Bonchi, C.~Castillo, and S.~Hajian,
  ``Fair top-k ranking with multiple protected groups,'' \emph{Information
  Processing \& Management}, vol.~59, no.~1, p. 102707, 2022.

\bibitem{dwork2012fairness}
C.~Dwork, M.~Hardt, T.~Pitassi, O.~Reingold, and R.~Zemel, ``Fairness through
  awareness,'' in \emph{Proceedings of the 3rd innovations in theoretical
  computer science conference}, 2012, pp. 214--226.

\bibitem{singh2018fairness}
A.~Singh and T.~Joachims, ``Fairness of exposure in rankings,'' in
  \emph{Proceedings of the 24th ACM SIGKDD International Conference on
  Knowledge Discovery \& Data Mining}, 2018, pp. 2219--2228.

\bibitem{gupta2021online}
A.~Gupta, E.~Johnson, J.~Payan, A.~K. Roy, A.~Kobren, S.~Panda, J.-B. Tristan,
  and M.~Wick, ``Online post-processing in rankings for fair utility
  maximization,'' in \emph{Proceedings of the 14th ACM International Conference
  on Web Search and Data Mining}, 2021, pp. 454--462.

\bibitem{kleinberg2016inherent}
J.~Kleinberg, S.~Mullainathan, and M.~Raghavan, ``Inherent trade-offs in the
  fair determination of risk scores,'' \emph{arXiv preprint arXiv:1609.05807},
  2016.

\bibitem{nycd1}
\BIBentryALTinterwordspacing
NYTimes, ``A {M}anhattan {D}istrict {W}here {S}chool {C}hoice {A}mounts to
  {S}egregation,'' 2017. [Online]. Available:
  \url{https://www.nytimes.com/2017/06/07/nyregion/a-manhattan-district-where-school-choice-amounts-to-segregation.html}
\BIBentrySTDinterwordspacing

\bibitem{collegelotteries2021}
\BIBentryALTinterwordspacing
D.~J. Baker and M.~N. Bastedo, ``What if we leave it up to chance? admissions
  lotteries and equitable access at selective colleges,'' \emph{Educational
  Researcher}, vol.~0, no.~0, p. 0013189X211055494, 2021. [Online]. Available:
  \url{https://doi.org/10.3102/0013189X211055494}
\BIBentrySTDinterwordspacing

\bibitem{ellison2021efficiency}
G.~Ellison and P.~A. Pathak, ``The efficiency of race-neutral alternatives to
  race-based affirmative action: Evidence from chicago's exam schools,''
  \emph{American Economic Review}, vol. 111, no.~3, pp. 943--75, 2021.

\bibitem{ehlers2014}
L.~Ehlers, I.~E. Hafalir, M.~B. Yenmez, and M.~A. Yildirim, ``School choice
  with controlled choice constraints: Hard bounds versus soft bounds,''
  \emph{Journal of Economic Theory}, vol. 153, pp. 648--683, 2014.

\bibitem{kojima2012}
F.~Kojima, ``School choice: Impossibilities for affirmative action,''
  \emph{Games and Economic Behavior}, vol.~75, no.~2, pp. 685--693, 2012.

\bibitem{hafalir2013}
I.~E. Hafalir, M.~B. Yenmez, and M.~A. Yildirim, ``Effective affirmative action
  in school choice,'' \emph{Theoretical Economics}, vol.~8, no.~2, pp.
  325--363, 2013.

\bibitem{aziz2021multi}
H.~Aziz and Z.~Sun, ``Multi-rank smart reserves,'' in \emph{Proceedings of the
  22nd ACM Conference on Economics and Computation}, 2021, pp. 105--124.

\bibitem{abdulkadirouglu2021priority}
A.~Abdulkadiro{\u{g}}lu and A.~Grigoryan, ``Priority-based assignment with
  reserves and quotas,'' National Bureau of Economic Research, Tech. Rep.,
  2021.

\bibitem{NYCmatching}
A.~Abdulkadiro{\u{g}}lu, P.~A. Pathak, and A.~E. Roth, ``The {N}ew {Y}ork
  {C}ity {H}igh {S}chool {M}atch,'' \emph{American Economic Review}, vol.~95,
  no.~2, pp. 364--367, 2005.

\bibitem{galeshapley}
D.~Gale and L.~S. Shapley, ``College admissions and the stability of
  marriage,'' \emph{The American Mathematical Monthly}, vol.~69, no.~1, pp.
  9--15, 1962.

\bibitem{affelnet}
G.~Fack and J.~Grenet, ``Mixité sociale et scolaire dans les lycées parisiens
  : Les enseignements de la procédure affelnet,'' 09 2016.

\bibitem{angwin_larson_2016}
\BIBentryALTinterwordspacing
J.~Angwin and J.~Larson, ``How we analyzed the compas recidivism algorithm,''
  May 2016. [Online]. Available:
  \url{https://www.propublica.org/article/how-we-analyzed-the-compas-recidivism-algorithm}
\BIBentrySTDinterwordspacing

\bibitem{Rudin2020Age}
\BIBentryALTinterwordspacing
C.~Rudin, C.~Wang, and B.~Coker, ``The age of secrecy and unfairness in
  recidivism prediction,'' \emph{Harvard Data Science Review}, vol.~2, no.~1, 3
  2020, https://hdsr.mitpress.mit.edu/pub/7z10o269. [Online]. Available:
  \url{https://hdsr.mitpress.mit.edu/pub/7z10o269}
\BIBentrySTDinterwordspacing

\bibitem{Jackson2020Setting}
\BIBentryALTinterwordspacing
E.~Jackson and C.~Mendoza, ``Setting the record straight: What the compas core
  risk and need assessment is and is not,'' \emph{Harvard Data Science Review},
  vol.~2, no.~1, 3 2020, https://hdsr.mitpress.mit.edu/pub/hzwo7ax4. [Online].
  Available: \url{https://hdsr.mitpress.mit.edu/pub/hzwo7ax4}
\BIBentrySTDinterwordspacing

\bibitem{COMPASmessy21}
\BIBentryALTinterwordspacing
M.~Bao, A.~Zhou, S.~Zottola, B.~Brubach, S.~Desmarais, A.~Horowitz, K.~Lum, and
  S.~Venkatasubramanian, ``It's compaslicated: The messy relationship between
  rai datasets and algorithmic fairness benchmarks,'' 2021. [Online].
  Available: \url{https://arxiv.org/abs/2106.05498}
\BIBentrySTDinterwordspacing

\bibitem{goeljustice}
J.~Jung, C.~Concannon, R.~Shroff, S.~Goel, and D.~G. Goldstein, ``Creating
  {S}imple {R}ules for {C}omplex {D}ecisions,'' \emph{Harvard {B}usiness
  {R}eview}, 2017,
  \url{https://hbr.org/2017/04/creating-simple-rules-for-complex-decisions}.

\bibitem{belkhir2017per}
N.~Belkhir, J.~Dr{\'e}o, P.~Sav{\'e}ant, and M.~Schoenauer, ``Per instance
  algorithm configuration of cma-es with limited budget,'' in \emph{Proceedings
  of the Genetic and Evolutionary Computation Conference}, 2017, pp. 681--688.

\bibitem{ruppert2011statistics}
D.~Ruppert and D.~S. Matteson, \emph{Statistics and data analysis for financial
  engineering}.\hskip 1em plus 0.5em minus 0.4em\relax Springer, 2011, vol.~13.

\bibitem{kingma2017adam}
D.~P. Kingma and J.~Ba, ``Adam: A method for stochastic optimization,'' 2017.

\bibitem{nycdata}
N.~DOE, ``Doing research in or about new york city public schools,'' 2022,
  \url{https://infohub.nyced.org/reports-and-policies/research/doing-research-in-new-york-city-public-schools}.

\bibitem{jarvelin2002cumulated}
K.~J{\"a}rvelin and J.~Kek{\"a}l{\"a}inen, ``Cumulated gain-based evaluation of
  ir techniques,'' \emph{ACM Transactions on Information Systems (TOIS)},
  vol.~20, no.~4, pp. 422--446, 2002.

\bibitem{zafar2017fairness}
M.~B. Zafar, I.~Valera, M.~G. Rogriguez, and K.~P. Gummadi, ``Fairness
  constraints: Mechanisms for fair classification,'' in \emph{Artificial
  intelligence and statistics}.\hskip 1em plus 0.5em minus 0.4em\relax PMLR,
  2017, pp. 962--970.

\end{thebibliography}

\end{document}